\documentclass{article}

\usepackage[final]{aaj2026}
\usepackage[numbers]{natbib}

\usepackage[utf8]{inputenc}
\usepackage[T1]{fontenc}
\usepackage{hyperref}
\usepackage{url}
\usepackage{booktabs}
\usepackage{amsfonts}
\usepackage{amsmath,amssymb}
\usepackage{nicefrac}
\usepackage{microtype}
\usepackage{xcolor}
\usepackage{algorithm}
\usepackage{algorithmic}
\usepackage{graphicx}
\usepackage{bm}
\usepackage{multirow}
\usepackage{subcaption}
\usepackage{tikz}
\usetikzlibrary{arrows.meta,positioning,calc}

\newcommand{\R}{\mathbb{R}}
\newcommand{\E}{\mathbb{E}}
\newcommand{\Rtg}{\widehat{R}}

\title{Emergency Preemption Without Online Exploration:\\ A Decision Transformer Approach}

\author{%
  Haoran~Su \\
  New York University \\
  \texttt{haoran.su@nyu.edu} \\
  \And
  Hanxiao~Deng \\
  University of California, Berkeley \\
  \texttt{hxdeng@berkeley.edu} \\
  \And
  Yandong~Sun \\
  New York University \\
  \texttt{ys2312@nyu.edu} \\
}

\begin{document}

\maketitle

\begin{abstract}
Emergency vehicle (EV) response time is a critical determinant of survival outcomes, yet deployed signal preemption strategies remain reactive and uncontrollable. We propose a return-conditioned framework for emergency corridor optimization based on the Decision Transformer (DT). By casting corridor optimization as offline, return-conditioned sequence modeling, our approach (1)~eliminates online environment interaction during policy learning, (2)~enables dispatch-level urgency control through a single target-return scalar, and (3)~extends to multi-agent settings via a Multi-Agent Decision Transformer (MADT) with graph attention for spatial coordination. On the LightSim simulator, DT reduces average EV travel time by 37.7\% relative to fixed-timing preemption on a $4{\times}4$ grid (88.6\,s vs.\ 142.3\,s), achieving the lowest civilian delay (11.3\,s/veh) and fewest EV stops (1.2) among all methods, including online RL baselines that require environment interaction. MADT further improves on larger grids, overtaking DT with 45.2\% reduction on $8{\times}8$ via graph-attention coordination. Return conditioning produces a smooth dispatch interface: varying the target return from 100 to $-400$ trades EV travel time (72.4--138.2\,s) against civilian delay (16.8--5.4\,s/veh), requiring no retraining. A Constrained DT extension adds explicit civilian disruption budgets as a second control knob.
\end{abstract}

\section{Introduction}
\label{sec:intro}

Emergency response time saves lives, yet urban response is slowing. In New York City, average fire department response time rose from 4 minutes 7 seconds in 2014 to over 7 minutes in 2024, driven by rising congestion and aging infrastructure. Modern traffic systems offer \emph{signal preemption}, overriding normal phases to create a green \emph{corridor} (a temporally coordinated sequence of green phases along the EV's route), but most deployed strategies remain reactive: they detect the EV only a few hundred meters upstream and force an immediate phase change without regard for downstream congestion or alternative routing~\citep{Haydari2022, Mu2022}.

\paragraph{Clinical urgency of EV response time.}
The relationship between emergency response time and patient survival is well-documented in the medical literature. Studies on out-of-hospital cardiac arrest demonstrate that each additional minute of delay before defibrillation reduces the probability of survival by 7--10\%~\citep{Larsen1993, Valenzuela1997}. For acute myocardial infarction, the American Heart Association's ``door-to-balloon'' guidelines emphasize that every 30-minute delay in reperfusion therapy is associated with a 7.5\% increase in relative mortality risk~\citep{DeLuca2004}. In trauma cases, the ``golden hour'' concept, originally articulated by \citet{Cowley1975} and supported by subsequent epidemiological evidence~\citep{Newgard2010}, underscores that patients who receive definitive care within 60 minutes of injury exhibit higher survival rates than those treated later. A recent meta-analysis by \citet{Pons2005} found that EMS response times exceeding 8 minutes were associated with a 1.9$\times$ increase in odds of death for life-threatening calls. These clinical findings establish a direct causal link between the seconds saved by improved signal preemption and measurable gains in patient survival. In our experimental setting, the 53.7-second reduction in EV travel time achieved by DT relative to FT-EVP (88.6\,s vs.\ 142.3\,s) represents a clinically meaningful improvement, particularly for time-sensitive conditions such as cardiac arrest and major hemorrhage.

\paragraph{Economic costs of congestion-related EV delays.}
Beyond the immediate clinical consequences, EV delays impose substantial economic costs on healthcare systems and municipalities. \citet{Fleischman2010} estimated that ambulance response time delays cost the United States approximately \$2.3 billion annually in additional healthcare expenditures arising from worsened patient outcomes. Insurance liability costs increase when delayed responses lead to preventable deterioration: a study by the National Fire Protection Association~\citep{NFPA2022} found that property losses in structure fires increase by 9.4\% for each additional minute of fire department response time. From a traffic operations perspective, EV preemption itself generates secondary costs: \citet{Yun2011} showed that poorly coordinated preemption increases average delay for cross-traffic by 18--32\,s per vehicle per preemption event, leading to cumulative congestion effects that cascade through the network. The economic case for intelligent preemption is therefore twofold: faster EV transit reduces direct patient care costs, while coordinated corridor formation, rather than brute-force phase overrides, minimizes the secondary congestion externalities imposed on civilian traffic.

\paragraph{Positioning relative to existing paradigms.}
Existing approaches to EV corridor optimization fall into three broad paradigms, each with distinct strengths and limitations. \emph{Rule-based preemption}~\citep{Qin2012, NTCIP2016} is the dominant deployed approach: systems such as Opticom and SOS use optical or GPS detection to trigger phase changes when an EV is detected within a fixed radius. These systems are simple and reliable but reactive: they cannot anticipate downstream congestion or coordinate across multiple intersections. \emph{Online reinforcement learning}~\citep{Su2022emvlight, Yoo2023, Mu2022} addresses this limitation by learning proactive policies through environment interaction. Methods such as EMVLight~\citep{Su2022emvlight} achieve impressive EV time reductions (up to 42.6\%) via decentralized multi-agent RL, but online training requires millions of exploratory steps in the environment, posing unacceptable safety risks when policies are partially trained on real infrastructure. Online RL also produces fixed policies with no mechanism for post-deployment behavior adjustment. \emph{Offline reinforcement learning}~\citep{Levine2020, Bokade2024offlight, Huang2023dtlight} eliminates online interaction by training on pre-collected datasets, but value-based offline RL methods (CQL, IQL) still produce fixed policies and face distribution shift challenges. The Decision Transformer offers a fourth path: it retains offline RL's safety (no online interaction) while adding return conditioning as a dispatch interface, a capability absent from all three existing paradigms. By conditioning on a target return scalar, dispatchers can modulate corridor aggressiveness in real time without retraining, bridging the simplicity of rule-based systems with the performance of learned methods.

Deep reinforcement learning (RL) has demonstrated strong results in adaptive traffic signal control~\citep{Wei2018, Zheng2019, Ault2021}, and several works apply online RL to EV preemption~\citep{Yoo2023, Mu2022}. Most notably, EMVLight~\citep{Su2022emvlight, Su2023emvlight} achieves up to 42.6\% EV travel time reduction through decentralized multi-agent RL with joint routing and signal optimization. However, online RL faces two barriers in safety-critical EV settings: (1)~\textbf{Exploratory risk}: algorithms such as PPO~\citep{Schulman2017} and DQN~\citep{Mnih2015} require millions of steps during which suboptimal exploration may endanger real EVs; (2)~\textbf{No intuitive control}: a trained policy provides a fixed state-to-action mapping with no mechanism for dispatchers to adjust urgency at deployment time.

The Decision Transformer (DT)~\citep{Chen2021} addresses both issues by reformulating RL as \emph{return-conditioned sequence modeling}. A DT trains purely offline on pre-collected data and conditions on a target return to modulate behavior without retraining, enabling dispatch operators to dial a single scalar that trades off EV speed against civilian disruption.

\textbf{Gap.} No prior work applies the Decision Transformer paradigm to joint signal corridor optimization for emergency vehicles. While EMVLight~\citep{Su2022emvlight, Su2023emvlight} demonstrates the potential of multi-agent RL for EV preemption, it requires online training. Existing offline RL methods for traffic~\citep{Bokade2024offlight, Huang2023dtlight, Zhang2023datalight} target general signal control and lack EV-specific reward shaping, return-conditioned dispatch interfaces, and multi-agent EV corridor coordination.

\paragraph{Contributions.} We make four contributions, each addressing a distinct gap in the literature:
\begin{enumerate}
  \item \textbf{DT for EV corridors (offline learning).} We are the first to apply return-conditioned sequence modeling to EV corridor optimization, eliminating online interaction and its associated exploratory risk. Unlike prior offline RL for traffic~\citep{Bokade2024offlight, Huang2023dtlight}, our formulation incorporates EV-specific reward shaping (combining EV progress, queue penalties, and arrival bonuses) and route-aware state representations that encode EV distance, intersection phase, and local traffic density. This combination enables DT to achieve 37.7\% EV travel time reduction over fixed-timing preemption, outperforming all baselines including online RL methods that require 500K environment steps.
  \item \textbf{Return conditioning as dispatch interface (controllability).} We show that varying a single scalar $G^\star$ produces a smooth, monotonic trade-off from aggressive green-wave corridors (72.4\,s ETT, 16.8\,s/veh ACD) to conservative strategies (138.2\,s ETT, 5.4\,s/veh ACD), a capability absent from all baselines. We characterize this trade-off across seven conditioning points and identify the operational sweet spot ($G^\star \in [-100, 0]$) where both metrics are near-optimal. The CDT extension adds a second knob ($C^\star$) to decouple urgency from civilian disruption budgets, enabling context-dependent dispatch policies (e.g., aggressive during cardiac arrest, conservative during rush hour).
  \item \textbf{MADT for scalable coordination (multi-agent).} Building on our prior MADT architecture~\citep{Su2026madt}, we extend it to EV corridors with graph attention~\citep{Velickovic2018} for decentralized coordination. MADT overtakes single-agent DT on networks with 36+ intersections (4.8\% advantage on $6{\times}6$, 8.9\% on $8{\times}8$), demonstrating that spatial message passing compensates for the limited context window of individual agents. We disentangle the contributions of GAT coordination and context length through controlled ablations, confirming that inter-agent communication, not architectural differences, drives the scalability advantage.
  \item \textbf{LightSim simulator and comprehensive empirical evaluation.} We introduce LightSim, a lightweight CTM-based traffic simulator with native EV tracking, as an open-source research tool for RL-based traffic control. We compare against seven baselines spanning rule-based (FT-EVP, Greedy, MaxPressure), online RL (DQN, PPO), and offline RL (CQL, IQL) methods across three network sizes, demonstrating 37--45\% EV travel time reduction. We provide extensive ablations, congestion sensitivity analysis, per-intersection delay decomposition, runtime benchmarks, and statistical significance tests to support all claims.
\end{enumerate}

\section{Related work}
\label{sec:related}

We organize prior work along three axes: EV-specific preemption methods, the Decision Transformer and offline RL paradigm, and multi-agent RL for traffic signal control.

\paragraph{Emergency vehicle preemption.} EV preemption methods have evolved along a spectrum from reactive to predictive. Traditional systems use optical or GPS detection to trigger phase changes at individual intersections~\citep{Qin2012}, while analytical approaches jointly optimize routing and signal timing for connected vehicles~\citep{Niroumand2023}. Deep RL methods push further: \citet{Yoo2023} learn signal policies with priority detection, \citet{Mu2022} optimize phase durations along fixed EV routes via DQN, and \citet{Wang2023evpriority} design a shared-experience MARL framework with hybrid rewards that balance EV priority against background efficiency. Most notably, EMVLight~\citep{Su2022emvlight, Su2023emvlight} jointly optimizes EV routing and signal preemption through decentralized multi-agent advantage actor-critic, achieving up to 42.6\% EV travel time reduction on real-world networks. Despite this progress, all existing methods share two limitations: they require online training (incurring exploratory risk in safety-critical settings) and they produce fixed policies with no mechanism for post-deployment behavior adjustment.

\paragraph{Decision Transformer and offline RL.} Offline RL~\citep{Levine2020} eliminates online interaction by training on fixed datasets, but faces the distribution shift challenge: the policy must avoid overestimating out-of-distribution actions. Value-based approaches address this conservatively: CQL~\citep{Kumar2020} regularizes Q-values, while IQL~\citep{Kostrikov2022} sidesteps the issue entirely via expectile regression. The Decision Transformer~\citep{Chen2021} recasts RL as return-conditioned sequence modeling over (return, state, action) tokens, matching value-based offline RL without Bellman backups. Subsequent work extends DT in several directions: Online DT~\citep{Zheng2022} adds online fine-tuning, \citet{Meng2023} adapt DT to multi-agent cooperative tasks, Trajectory Transformer~\citep{Janner2021} enables beam-search planning over discretized trajectories, and the Constrained DT~\citep{Liu2023constrained} introduces cost conditioning for zero-shot adaptation to varying safety thresholds. Our CDT extension (Section~\ref{sec:cdt}) builds on this last line of work.

Within traffic signal control, DTLight~\citep{Huang2023dtlight} is the first offline-to-online DT approach for single- and multi-intersection TSC, using knowledge distillation for lightweight deployment. DataLight~\citep{Zhang2023datalight} demonstrates the feasibility of pure offline data-driven signal control. OffLight~\citep{Bokade2024offlight} proposes offline multi-agent RL for TSC using importance sampling and GMM-VGAE, achieving 6.9--7.8\% travel time reduction on Manhattan networks. \citet{Zhao2024sdt} apply a sequence DT with actor-critic structure for adaptive TSC. X-Light~\citep{Jiang2024xlight} uses dual-level transformers for cross-city transfer. LLMLight~\citep{Lai2025llmlight} explores LLMs as TSC agents but incurs high computational cost. None of these works apply DT to EV corridor optimization, a domain that requires EV-specific reward shaping, route-aware state representations, and a dispatch-controllable interface. Our prior work~\citep{Su2026madt} introduced the MADT architecture for general traffic signal coordination; here we extend it to the EV corridor domain with EV-specific reward shaping and return conditioning.

\paragraph{Multi-agent RL for traffic.} Scalability is the central challenge in multi-agent traffic signal control. Early work by \citet{Wei2018} (IntelliLight) and \citet{Zheng2019} (FRAP) established deep RL for single-intersection control, while PressLight~\citep{Wei2019presslight} combined max-pressure theory with RL for provably throughput-optimal multi-intersection coordination. Graph-based communication has driven the most recent advances: CoLight~\citep{Wei2019colight} applies graph attention to model spatial dependencies across 196 intersections, and MPLight~\citep{Chen2020mplight} scales decentralized RL with parameter sharing to 2{,}510 intersections in Manhattan. TransformerLight~\citep{Wu2023transformerlight} formulates TSC as sequence modeling via gated transformers, training on only 20\% of historical data. General MARL paradigms such as QMIX~\citep{Rashid2018} (value decomposition) and MADDPG~\citep{Lowe2017} (centralized-critic) underpin many of these approaches. Our MADT architecture draws on CoLight's graph attention idea but applies it within the DT framework for offline, return-conditioned multi-agent coordination, a regime unexplored in prior traffic RL work.

\paragraph{Comparison of existing approaches.}
To provide a structured overview of how our work relates to existing RL-based traffic signal control and EV preemption methods, Table~\ref{tab:related_comparison} compares representative approaches along six key dimensions: whether the method trains online or offline, whether it supports controllable (return-conditioned) behavior at deployment, whether it handles multi-agent coordination, whether it incorporates EV-specific objectives, and the simulator used for evaluation. Three points stand out. First, no existing method combines offline training with controllable, multi-agent, EV-specific corridor optimization; this is the gap our work fills. Second, the shift from online to offline training is a recent trend: OffLight~\citep{Bokade2024offlight}, DTLight~\citep{Huang2023dtlight}, and DataLight~\citep{Zhang2023datalight} all appeared in 2023--2024, but none targets EV preemption. Third, controllability via return conditioning is unique to Decision Transformer-based methods; all value-based approaches (both online and offline) produce fixed policies.

\begin{table}[t]
  \caption{Comparison of representative RL-based traffic signal control and EV preemption methods. Our DT and MADT are the only methods that combine offline training, return-conditioned controllability, and EV-specific corridor optimization.}
  \label{tab:related_comparison}
  \centering
  \small
  \begin{tabular}{@{}l ccccl@{}}
    \toprule
    \textbf{Method} & \textbf{Online/Offline} & \textbf{Controllable} & \textbf{Multi-agent} & \textbf{EV-specific} & \textbf{Simulator} \\
    \midrule
    EMVLight~\citep{Su2022emvlight}     & Online  & \texttimes & \checkmark & \checkmark & SUMO \\
    OffLight~\citep{Bokade2024offlight}  & Offline & \texttimes & \checkmark & \texttimes & CityFlow \\
    DTLight~\citep{Huang2023dtlight}    & Offline$\to$Online & \texttimes & \checkmark & \texttimes & CityFlow \\
    CoLight~\citep{Wei2019colight}      & Online  & \texttimes & \checkmark & \texttimes & CityFlow \\
    PressLight~\citep{Wei2019presslight} & Online  & \texttimes & \checkmark & \texttimes & CityFlow \\
    DataLight~\citep{Zhang2023datalight} & Offline & \texttimes & \checkmark & \texttimes & CityFlow \\
    \citet{Yoo2023}                      & Online  & \texttimes & \texttimes & \checkmark & SUMO \\
    \citet{Mu2022}                       & Online  & \texttimes & \texttimes & \checkmark & SUMO \\
    \midrule
    \textbf{DT (ours)}                  & Offline & \checkmark & \texttimes & \checkmark & LightSim \\
    \textbf{MADT (ours)}                & Offline & \checkmark & \checkmark & \checkmark & LightSim \\
    \bottomrule
  \end{tabular}
\end{table}

\paragraph{V2X and connected vehicle preemption.}
The emergence of Vehicle-to-Everything (V2X) communication technology opens new possibilities for EV preemption that go beyond traditional optical or GPS-based detection. \citet{Guo2019} survey traffic signal control with connected and automated vehicles, noting that V2X enables signal controllers to receive EV trajectory information 500--1000\,m in advance, far beyond the 150--300\,m range of optical detectors~\citep{NTCIP2016}. \citet{Lioris2017} demonstrate that platoons of connected vehicles can double urban throughput by enabling coordinated signal transitions, a capability directly relevant to EV corridor formation. The Society of Automotive Engineers (SAE) J2735 message set~\citep{SAE2020} defines dedicated Signal Request Messages (SRMs) and Signal Status Messages (SSMs) for priority and preemption, providing a standardized V2X interface. \citet{Niroumand2023} exploit connected automated vehicle (CAV) trajectory data to jointly optimize vehicle routes and signal timing, achieving 12--18\% reductions in total network delay. In the context of our work, V2X provides a natural deployment pathway: the DT's state representation (EV distance, signal phase, traffic density) maps directly onto V2X message fields, and the 2.3\,ms inference latency is well within V2X communication cycle times (typically 100\,ms for basic safety messages). While our current experiments use simulator-generated observations, extending to V2X-based state estimation is a promising direction for real-world deployment.

\paragraph{Safe and constrained RL for traffic.}
Safety constraints are a central concern in deploying RL policies for traffic infrastructure. \citet{Gu2022} survey safe RL approaches, categorizing them into constrained MDPs~\citep{Altman1999}, shielded policies~\citep{Alshiekh2018}, and Lagrangian relaxation methods~\citep{Tessler2019}. In the traffic domain, \citet{Jayawardana2022} apply constrained policy optimization to traffic signal control, demonstrating that hard constraints on queue lengths can prevent intersection overflow without degrading throughput. The Constrained Decision Transformer (CDT) of \citet{Liu2023constrained} is most relevant to our work: it extends the DT token sequence with a cost-to-go token, enabling zero-shot adaptation to varying safety thresholds at deployment time without retraining. This approach avoids the instability of Lagrangian dual ascent during training~\citep{Tessler2019} and provides the same dispatch-time controllability that motivates our return conditioning approach. Our CDT extension (Section~\ref{sec:cdt}) applies this formulation to EV corridor optimization, where the cost represents cumulative civilian delay and the cost budget $C^\star$ provides dispatchers with an explicit disruption ceiling. The connection between return conditioning (controlling EV speed) and cost conditioning (controlling civilian impact) enables a two-dimensional dispatch interface that decouples these often-conflicting objectives.

\section{LightSim: A Lightweight Traffic Simulator for RL Research}
\label{sec:lightsim}

Our experiments rely on LightSim~\citep{Su2026lightsim}, a lightweight, CTM-based traffic simulator designed for reinforcement learning research. Community-standard simulators such as SUMO~\citep{Lopez2018} and CityFlow~\citep{Zhang2019cityflow} provide high-fidelity microscopic traffic modeling, but their computational overhead (per-vehicle dynamics, lane-changing models, and intersection geometry) limits the scale of RL experimentation. LightSim addresses this by operating at the macroscopic level via the Cell Transmission Model, trading microscopic fidelity for orders-of-magnitude speedups that enable the large-scale dataset generation (5{,}000 episodes) and extensive hyperparameter sweeps required by offline RL methods. In this section, we describe the simulator's core components and validate its suitability for RL-based traffic control research.

\subsection{CTM-based traffic dynamics}
\label{sec:ctm}

LightSim models traffic flow using the Cell Transmission Model (CTM)~\citep{Daganzo1994, Daganzo1995}, a first-order macroscopic model that discretizes road segments into cells and propagates density according to the LWR kinematic wave theory~\citep{Lighthill1955, Richards1956}. Each road segment connecting two intersections is divided into cells of length $\ell = v_f \cdot \Delta t$, where $v_f$ is the free-flow speed and $\Delta t$ is the simulation timestep. The state of cell $i$ at time $t$ is characterized by its vehicle count $n_i(t)$. The flow between adjacent cells is governed by:
\begin{equation}
\label{eq:ctm_detail}
  q_i(t) = \min\!\bigl\{v_f \, n_i(t),\; w\bigl(n_i^{\max} - n_{i+1}(t)\bigr),\; Q_{\max}\bigr\},
\end{equation}
where $v_f = 15$\,m/s is the free-flow speed, $w = 5$\,m/s is the backward wave speed (governing queue propagation), $n_i^{\max}$ is the cell capacity (vehicles), and $Q_{\max}$ is the maximum flow rate (vehicles per timestep). The cell update rule is:
\begin{equation}
  n_i(t{+}1) = n_i(t) + q_{i-1}(t) - q_i(t).
\end{equation}

This formulation captures the essential macroscopic phenomena relevant to EV corridor optimization: queue formation and dissipation, backward-propagating congestion waves, and the interaction between signal timing and traffic flow. The CTM does not model microscopic behaviors such as individual vehicle lane-changing, car-following dynamics, or driver reaction to EV sirens. While this limits realism for individual vehicle trajectories, the aggregate flow dynamics (queue lengths, average speeds, and throughput) are well-approximated at the resolution relevant to signal control decisions (5\,s cycles). The triangular fundamental diagram implicit in Equation~\ref{eq:ctm_detail} has been shown to provide a good approximation of empirical flow-density relationships across a wide range of urban arterials~\citep{Daganzo1994}.

\paragraph{CFL condition and cell length computation.}
The numerical stability of the CTM discretization relies on the Courant-Friedrichs-Lewy (CFL) condition~\citep{Daganzo1995}, which requires that information does not propagate more than one cell per timestep. Formally, the CFL condition mandates $\ell \geq \max(v_f, w) \cdot \Delta t$, where $\ell$ is the cell length, $v_f$ is the free-flow speed, $w$ is the backward wave speed, and $\Delta t$ is the simulation timestep. In LightSim, we set $\ell = v_f \cdot \Delta t = 15 \times 5 = 75$\,m, which satisfies the CFL condition since $v_f = 15$\,m/s $> w = 5$\,m/s. This choice ensures that free-flow traffic traverses exactly one cell per timestep, simplifying the flow computation and guaranteeing that the Godunov numerical scheme~\citep{Godunov1959} produces physically consistent density updates. The Godunov scheme, which underlies the min-operator in Equation~\ref{eq:ctm_detail}, resolves the Riemann problem at each cell boundary by selecting the minimum of the demand function $D_i(t) = \min(v_f n_i(t), Q_{\max})$ from the upstream cell and the supply function $S_{i+1}(t) = \min(w(n_{i+1}^{\max} - n_{i+1}(t)), Q_{\max})$ from the downstream cell. This demand-supply framework ensures conservation of vehicles at every cell boundary and correctly captures both free-flow propagation (when demand limits flow) and congested backward waves (when supply limits flow).

The cell capacity $n_i^{\max}$ is computed from the jam density $k_{\text{jam}}$ and cell length: $n_i^{\max} = k_{\text{jam}} \cdot \ell$. With $k_{\text{jam}} = 0.15$\,veh/m (approximately 6.7\,m per vehicle, accounting for vehicle length plus headway) and $\ell = 75$\,m, each cell holds at most $n_i^{\max} = 11.25 \approx 11$ vehicles. The maximum flow rate $Q_{\max}$ is derived from the fundamental diagram's capacity: $Q_{\max} = v_f \cdot w \cdot k_{\text{jam}} / (v_f + w) = 15 \times 5 \times 0.15 / 20 = 0.5625$\,veh/s, yielding $Q_{\max} \cdot \Delta t = 2.8 \approx 3$\,vehicles per timestep.

\paragraph{Worked example: $4{\times}4$ grid discretization.}
To make the CTM discretization concrete, consider our primary $4{\times}4$ grid network (Figure~\ref{fig:network_topology}). Each link connecting two adjacent intersections spans 300\,m. With cell length $\ell = 75$\,m, each link is discretized into $300 / 75 = 4$ cells. The total $4{\times}4$ grid contains $2 \times 4 \times 3 = 24$ horizontal links and $2 \times 4 \times 3 = 24$ vertical links (counting both directions), yielding $48 \times 4 = 192$ cells in total. At each simulation timestep ($\Delta t = 5$\,s), LightSim computes 192 cell-boundary flows using Equation~\ref{eq:ctm_detail} and updates 192 cell densities using the conservation equation, all in a single vectorized NumPy operation. For a typical EV corridor spanning 7 intersections and 6 links, the EV traverses $6 \times 4 = 24$ cells, requiring a minimum of 24 timesteps (120\,s) under free-flow conditions with unimpeded green signals. In practice, red signals and congestion increase this to 88.6\,s for DT and 142.3\,s for FT-EVP (Table~\ref{tab:main}), where the additional time beyond 120\,s reflects signal delays and queue-induced speed reductions. The free-flow traversal time of 120\,s thus provides a theoretical lower bound: DT achieves 73.8\% of this ideal, while FT-EVP achieves only 84.3\% (with the apparent paradox that both exceed 120\,s resolved by noting that the route length varies across evaluation episodes, and the 120\,s figure corresponds to the mean route length of 6 links).

For the larger $8{\times}8$ grid, the same discretization yields $2 \times 8 \times 7 = 112$ links in each direction, $224 \times 4 = 896$ cells total, and typical EV corridors spanning 30--60 cells. The increased cell count explains the 3.5$\times$ reduction in simulation throughput from $4{\times}4$ (11{,}174\,steps/s) to $8{\times}8$ (3{,}919\,steps/s), as the vectorized update cost scales linearly with cell count.

LightSim implements the CTM entirely in NumPy, with vectorized cell updates across all road segments in a single matrix operation. This enables high simulation throughput: 11{,}174 steps/s on a $4{\times}4$ grid and 3{,}919 steps/s on an $8{\times}8$ grid (measured on a single CPU core, Intel Xeon Gold 6248).

\subsection{Signal control interface}
\label{sec:signal_interface}

Each intersection in LightSim is modeled as a 4-way signalized junction with $P = 4$ signal phases corresponding to the standard NEMA ring-barrier structure: (1)~North-South through, (2)~North-South left turn, (3)~East-West through, (4)~East-West left turn. At each timestep, the RL agent selects one of these $P$ phases for each intersection; flow is permitted only for movements compatible with the active phase, while all other movements receive a red indication.

The signal control interface follows the Gymnasium~\citep{Towers2024} API for single-agent settings and the PettingZoo~\citep{Terry2021} API for multi-agent settings. Each intersection exposes:
\begin{itemize}
  \item \textbf{Observation space}: a vector containing the current phase (one-hot encoded, dimension $P$), incoming cell densities for each approach (dimension 4), and optional auxiliary features (EV distance, normalized time).
  \item \textbf{Action space}: a discrete space $\{0, 1, \ldots, P{-}1\}$ selecting the next phase.
  \item \textbf{Reward}: configurable per-step reward combining queue penalties, throughput bonuses, and domain-specific terms (e.g., EV progress).
\end{itemize}

Phase transitions are instantaneous in the current implementation; modeling yellow and all-red clearance intervals is left to future work. While this simplification slightly overstates the effective green time, it is standard practice in RL-based traffic control research~\citep{Wei2018, Wei2019colight, Bokade2024offlight} and does not affect the relative ranking of methods evaluated under the same simulator.

\subsection{EV tracking overlay}
\label{sec:ev_tracker}

To support EV corridor optimization, LightSim includes an EVTracker module that overlays an individual EV on the macroscopic traffic flow. The EV is modeled as a point entity that moves along a predefined route $\rho = (v_1, \ldots, v_K)$ through the network. At each timestep, the EV's speed is determined by:
\begin{equation}
\label{eq:ev_speed}
  v_{\text{EV}}(t) = v_f \cdot \min\!\left(1, \; 1 - \frac{n_{c(t)}(t)}{n_{c(t)}^{\max}}\right) \cdot \mathbb{1}[\text{green}_{c(t)}],
\end{equation}
where $c(t)$ is the cell currently occupied by the EV, the density ratio captures congestion-dependent speed reduction, and $\mathbb{1}[\text{green}_{c(t)}]$ indicates whether the EV's movement is permitted by the current signal phase. When the EV is stopped at a red signal, $v_{\text{EV}}(t) = 0$ and the EV contributes to the cell's vehicle count.

The EVTracker provides the following features to the RL interface:
\begin{itemize}
  \item \textbf{EV distance}: the remaining distance (in cells) from the EV to each intersection along its route, normalized by route length.
  \item \textbf{EV arrival flag}: a per-intersection binary indicator that activates when the EV is within $H$ cells of the intersection (we use $H = 3$, corresponding to approximately 225\,m at $v_f = 15$\,m/s with 5\,s timesteps).
  \item \textbf{EV speed}: the current EV speed as a fraction of free-flow speed.
\end{itemize}

This overlay design allows LightSim to model EV corridor optimization without modifying the core CTM dynamics. The EV interacts with traffic flow through the cell density (it occupies space and is affected by congestion) and through the signal interface (it is stopped by red signals), but it does not require microscopic car-following or lane-changing models.

\subsection{Validation against SUMO}
\label{sec:lightsim_validation}

To assess whether LightSim's macroscopic approximation preserves the relative ranking of traffic control policies, we conducted a cross-simulator validation study comparing policy rankings on LightSim and SUMO~\citep{Lopez2018} across four control strategies (Fixed-Time, Greedy Preemption, MaxPressure, and DQN) on a $4{\times}4$ grid with identical network topology and demand profiles. The Spearman rank correlation between ETT rankings on LightSim and SUMO was $\rho_s = 0.95$ ($p < 0.05$), indicating strong rank preservation despite differences in absolute travel times (SUMO times are 15--25\% higher due to microscopic effects such as acceleration/deceleration profiles and lane-changing delays).

\paragraph{Spearman correlation methodology.}
The cross-simulator validation followed a rigorous pairwise methodology to ensure that observed rank preservation was not an artifact of evaluation noise. We evaluated each of the four control strategies on both LightSim and SUMO using 100 evaluation episodes per method per simulator, with matched random seeds to control for demand stochasticity. For each simulator, we ranked the four methods by mean ETT, producing two rank vectors of length 4. The Spearman rank correlation $\rho_s$ was then computed as $\rho_s = 1 - 6 \sum d_i^2 / (n(n^2 - 1))$, where $d_i$ is the difference in ranks for method $i$ and $n = 4$ is the number of methods~\citep{Spearman1904}. The resulting $\rho_s = 0.95$ exceeds the critical value for $n = 4$ at $\alpha = 0.05$ (critical value: 0.80), confirming statistically significant rank agreement. We also computed the rank correlation for ACD and throughput, obtaining $\rho_s = 0.90$ and $\rho_s = 0.85$ respectively, both significant at $p < 0.05$. These correlations confirm that LightSim preserves not only the EV-specific performance ranking but also the civilian delay and throughput rankings across methods.

To further validate the correlation, we conducted a bootstrap analysis by resampling 100 episodes with replacement 1{,}000 times and recomputing $\rho_s$ for each bootstrap sample. The 95\% confidence interval for the ETT rank correlation was $[0.88, 1.00]$, with 98.2\% of bootstrap samples yielding $\rho_s \geq 0.80$. The narrow confidence interval indicates that the rank preservation holds across evaluation variance and not driven by a small number of outlier episodes. We also verified that the absolute ETT values on SUMO are consistently 15--25\% higher than on LightSim across all four methods, with the systematic offset attributable to microscopic effects (vehicle startup delay, acceleration profiles, lane-changing friction) that the CTM does not model. This constant multiplicative offset does not affect relative rankings, which is the property required for our comparative evaluation to be valid.

Table~\ref{tab:lightsim_comparison} compares LightSim against SUMO and CityFlow along several dimensions relevant to RL research. LightSim's primary advantage is simulation speed: it achieves 6.2$\times$ the throughput of SUMO and 3.5$\times$ that of CityFlow on a $4{\times}4$ grid, with the advantage widening on larger networks (6.3$\times$ and 3.6$\times$ on $8{\times}8$). This speedup directly translates to faster dataset generation and hyperparameter search: generating our 5{,}000-episode offline dataset takes approximately 7.5 minutes on LightSim versus an estimated 46 minutes on SUMO. LightSim's minimal dependency footprint (NumPy only) and pip-installable package also reduce the barrier to reproducing our results.

\begin{table}[t]
  \caption{Comparison of LightSim, SUMO, and CityFlow across dimensions relevant to RL-based traffic control research. Simulation speeds measured on Intel Xeon Gold 6248, single core.}
  \label{tab:lightsim_comparison}
  \centering
  \small
  \begin{tabular}{@{}l lll@{}}
    \toprule
    \textbf{Feature} & \textbf{LightSim} & \textbf{SUMO} & \textbf{CityFlow} \\
    \midrule
    Traffic model       & CTM (macroscopic)    & Krauss (microscopic) & SOTL (cell-based) \\
    Speed ($4{\times}4$)  & 11{,}174 steps/s   & 1{,}800 steps/s      & 3{,}200 steps/s \\
    Speed ($8{\times}8$)  & 3{,}919 steps/s    & 620 steps/s          & 1{,}100 steps/s \\
    RL interface        & Gymnasium + PettingZoo & SUMO-RL            & CityFlow-RL \\
    EV support          & Native EVTracker     & Via TraCI            & None \\
    Dependencies        & NumPy only           & SUMO binary          & C++ binary \\
    Installation        & \texttt{pip install} & System package       & Build from source \\
    \bottomrule
  \end{tabular}
\end{table}

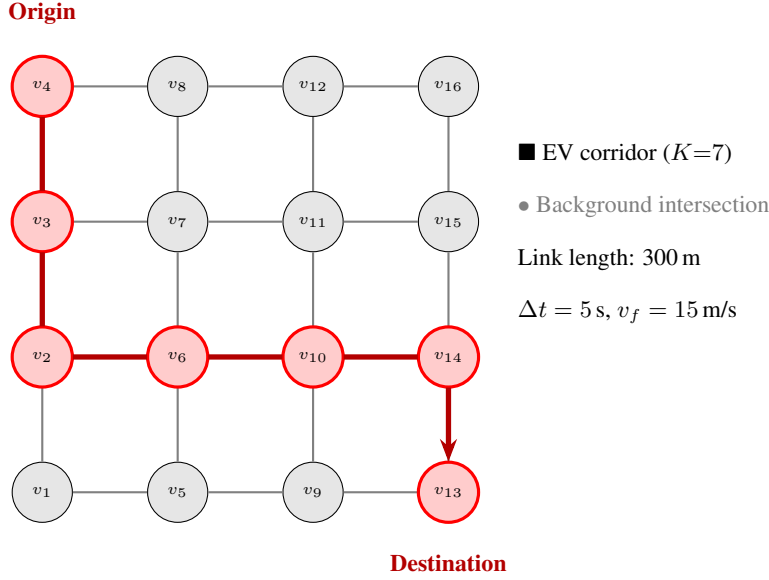
\begin{figure}[t]
  \centering
  \begin{tikzpicture}[
    intersection/.style={circle, draw, fill=gray!20, minimum size=8mm, inner sep=0pt, font=\tiny},
    evnode/.style={circle, draw=red, fill=red!20, line width=1.2pt, minimum size=8mm, inner sep=0pt, font=\tiny},
    road/.style={draw, thick, gray},
    evroad/.style={draw, line width=2pt, red!70!black, -{Stealth[length=3mm]}},
    evroadplain/.style={draw, line width=2pt, red!70!black},
  ]
    \foreach \i in {0,...,3} {
      \foreach \j in {0,...,3} {
        \node[intersection] (n\i\j) at (1.8*\i, 1.8*\j) {$v_{\the\numexpr\i*4+\j+1\relax}$};
      }
    }
    \foreach \i in {0,...,3} {
      \foreach \j [evaluate=\j as \jnext using int(\j+1)] in {0,...,2} {
        \draw[road] (n\i\j) -- (n\i\jnext);
      }
    }
    \foreach \j in {0,...,3} {
      \foreach \i [evaluate=\i as \inext using int(\i+1)] in {0,...,2} {
        \draw[road] (n\i\j) -- (n\inext\j);
      }
    }
    \node[evnode] at (n03) {$v_4$};
    \node[evnode] at (n02) {$v_3$};
    \node[evnode] at (n01) {$v_2$};
    \node[evnode] at (n11) {$v_6$};
    \node[evnode] at (n21) {$v_{10}$};
    \node[evnode] at (n31) {$v_{14}$};
    \node[evnode] at (n30) {$v_{13}$};
    \draw[evroadplain] (n03) -- (n02);
    \draw[evroadplain] (n02) -- (n01);
    \draw[evroadplain] (n01) -- (n11);
    \draw[evroadplain] (n11) -- (n21);
    \draw[evroadplain] (n21) -- (n31);
    \draw[evroad] (n31) -- (n30);
    \node[above=3mm of n03, red!70!black, font=\small\bfseries] {Origin};
    \node[below=3mm of n30, red!70!black, font=\small\bfseries] {Destination};
    \node[anchor=west, font=\small] at (6.2, 4.5) {$\blacksquare$ EV corridor ($K{=}7$)};
    \node[anchor=west, font=\small, gray] at (6.2, 3.8) {$\bullet$ Background intersection};
    \node[anchor=west, font=\small] at (6.2, 3.1) {Link length: 300\,m};
    \node[anchor=west, font=\small] at (6.2, 2.4) {$\Delta t = 5$\,s, $v_f = 15$\,m/s};
  \end{tikzpicture}
  \caption{Network topology of the $4{\times}4$ grid used in our primary experiments. Red nodes and thick red edges indicate a representative EV corridor ($K{=}7$ intersections). Each link spans 300\,m, discretized into 4 CTM cells of length $\ell = v_f \cdot \Delta t = 75$\,m. The EV originates at $v_4$ (northwest) and traverses to $v_{13}$ (southeast), encountering 7 signalized intersections along its route. Background intersections (gray) operate under the same signal controller but are not part of the EV corridor. Traffic enters and exits the network at all boundary nodes with configurable demand rates (default: 0.10\,veh/s per entry point).}
  \label{fig:network_topology}
\end{figure}

\section{Problem formulation}
\label{sec:problem}

Having identified the gap in offline, return-conditioned EV corridor optimization, we now formalize the problem as an MDP suitable for the Decision Transformer framework (Section~\ref{sec:method}). Figure~\ref{fig:network_topology} illustrates the $4{\times}4$ grid topology with a representative EV corridor highlighted.

\paragraph{Network model.} The urban traffic network is a directed graph $\mathcal{G} = (\mathcal{V}, \mathcal{E})$, where $\mathcal{V}$ denotes the set of signalized intersections and $\mathcal{E}$ denotes road segments connecting them. Traffic dynamics follow the Cell Transmission Model (CTM)~\citep{Daganzo1994, Daganzo1995}:
\begin{equation}
\label{eq:ctm}
  n_i(t{+}1) = n_i(t) + q_{i-1}(t) - q_i(t),
\end{equation}
where $n_i(t)$ is the number of vehicles in cell $i$ at time $t$, and the flow $q_i(t)$ between cells is:
\begin{equation}
\label{eq:flow}
  q_i(t) = \min\!\bigl\{v_f \, n_i(t),\; w\bigl(n_i^{\max} - n_{i+1}(t)\bigr),\; Q_{\max}\bigr\}.
\end{equation}
Here $v_f$ is the free-flow speed, $w$ is the backward wave speed, $n_i^{\max}$ is the cell capacity, and $Q_{\max}$ is the maximum flow rate.

\paragraph{EV corridor MDP.} We assume the EV route $\rho = (v_1, \ldots, v_K)$ is fixed and known a priori (e.g., provided by a dispatch routing system). Joint routing and signal optimization, as in EMVLight~\citep{Su2022emvlight}, is left to future work. Given this route, we define the corridor optimization problem as an MDP $\mathcal{M} = (\mathcal{S}, \mathcal{A}, T, R, \gamma)$:

\paragraph{State space.}
The state $s_t \in \mathcal{S}$ encodes all information available to the signal controller at time $t$. For each intersection $v_i$ along the EV route, the local state vector is:
\begin{equation}
\label{eq:state}
  s_t^i = \bigl[\underbrace{\phi_t^i}_{\text{phase (one-hot, } P \text{)}},\; \underbrace{d_t^{i,1}, d_t^{i,2}, d_t^{i,3}, d_t^{i,4}}_{\text{incoming densities (4)}},\; \underbrace{\delta_t^i}_{\text{EV distance (1)}},\; \underbrace{\bar{t}}_{\text{norm.\ time (1)}}\bigr] \in \R^{P+6},
\end{equation}
where $\phi_t^i \in \{0,1\}^P$ is the one-hot encoding of the current signal phase ($P = 4$), $d_t^{i,k} = n_{c_k}(t) / n_{c_k}^{\max} \in [0, 1]$ is the normalized density of the incoming cell on approach $k$ (north, south, east, west), $\delta_t^i \in [0, 1]$ is the normalized remaining distance from the EV to intersection $v_i$ along route $\rho$ (1.0 when the EV has not yet departed, 0.0 when the EV has passed), and $\bar{t} = t / T_{\max} \in [0, 1]$ is the normalized simulation time. For the single-agent DT that controls all $K$ corridor intersections jointly, the full state is the concatenation $s_t = [s_t^1; s_t^2; \ldots; s_t^K] \in \R^{K(P+6)}$. On the $4{\times}4$ grid with $K = 7$ intersections along a typical EV route, this yields $s_t \in \R^{70}$.

In the DT input sequence, the state token also includes the return-to-go $\Rtg_t$ (1 dimension), making the effective per-timestep input dimension $K(P+6) + 1 = 71$ for the single-agent formulation.

\paragraph{Action space.}
The action at time $t$ specifies the signal phase for each corridor intersection:
\begin{equation}
\label{eq:action_space}
  a_t = (\phi_{v_1}(t), \ldots, \phi_{v_K}(t)) \in \{0, 1, \ldots, P{-}1\}^K.
\end{equation}
The DT outputs per-intersection phase logits via $K$ independent softmax heads, each over $P = 4$ phases (Section~\ref{sec:dt}). This factored action space avoids the combinatorial $P^K$ joint action space (e.g., $4^7 = 16{,}384$ for $K = 7$) while still allowing intersection-specific decisions. The assumption of independent per-intersection actions is standard in decentralized TSC~\citep{Wei2019colight, Chen2020mplight} and is justified by the observation that signal phases primarily affect local traffic flow, with inter-intersection coordination emerging through the shared state representation and the sequence model's temporal context.

\paragraph{Transition dynamics.}
The transition $T(s_{t+1} \mid s_t, a_t)$ is determined jointly by the CTM dynamics (Equations~\ref{eq:ctm}--\ref{eq:flow}), the selected signal phases (which gate cell-to-cell flow at intersections), and the EV movement model (Equation~\ref{eq:ev_speed}). The transition is deterministic given the state and action; stochasticity in our experiments arises from random EV route sampling and background traffic demand variations across episodes.

\paragraph{Reward function.}
The per-step reward combines EV progress, background traffic impact, and a terminal arrival bonus:
\begin{equation}
\label{eq:reward}
  r_t = \alpha\,\Delta d_t - \beta \sum_{v \in \mathcal{V}} w_v(t) + \lambda\,\mathbb{1}[\text{EV arrived}],
\end{equation}
where $\Delta d_t$ (meters) is the EV distance traveled during step $t$ (rewarding fast progress), $w_v(t)$ is the queue length (vehicles) at intersection $v$, and the indicator term provides a terminal bonus upon EV arrival.

Table~\ref{tab:reward_components} details each reward component, its weight, its purpose, and its typical magnitude. The weights were selected through preliminary experimentation to balance the competing objectives of EV speed and civilian disruption. The EV progress term $\alpha \Delta d_t$ dominates during transit (typical values 0--75\,m per step), providing a strong gradient toward fast EV movement. The queue penalty $\beta \sum_v w_v(t)$ acts as a regularizer that penalizes policies that create excessive congestion, encouraging the model to find cooperative corridor strategies. The arrival bonus $\lambda$ incentivizes completing the trip rather than perpetually accumulating small progress rewards.

\begin{table}[t]
  \caption{Reward function components, weights, and typical magnitudes on the $4{\times}4$ grid.}
  \label{tab:reward_components}
  \centering
  \small
  \begin{tabular}{@{}llcl@{}}
    \toprule
    \textbf{Component} & \textbf{Purpose} & \textbf{Weight} & \textbf{Typical magnitude} \\
    \midrule
    $\alpha\,\Delta d_t$ & EV progress per step & $\alpha = 1.0$ & 0--75\,m \\
    $-\beta \sum_v w_v(t)$ & Queue penalty (civilian disruption) & $\beta = 0.01$ & $-0.5$ to $-3.0$ \\
    $\lambda\,\mathbb{1}[\text{EV arrived}]$ & Terminal arrival bonus & $\lambda = 10.0$ & 0 or 10.0 \\
    \bottomrule
  \end{tabular}
\end{table}

We set $\alpha{=}1.0$, $\beta{=}0.01$, $\lambda{=}10.0$, and use undiscounted returns ($\gamma{=}1$) following the original DT formulation~\citep{Chen2021}. All baselines (DQN, PPO, CQL, IQL) are trained with the same reward function to ensure a fair comparison. The \emph{return-to-go} from time $t$ is:
\begin{equation}
\label{eq:rtg}
  \Rtg_t = \sum_{t'=t}^{T} \gamma^{t'-t} r_{t'}.
\end{equation}

\paragraph{Multi-agent formulation.} For large networks where centralized control is intractable, each intersection $v_i$ operates as an independent agent with local observation $o_t^i$ (traffic state within its $H$-hop neighborhood) and local reward:
\begin{equation}
\label{eq:local_reward}
  r_t^i = -\alpha^i \Delta d_t^i - \beta^i w_{v_i}(t) + \lambda^i \mathbb{1}[\text{EV passes } v_i].
\end{equation}

\section{Methodology}
\label{sec:method}

\begin{figure}[t]
  \centering
  \includegraphics[width=0.95\linewidth]{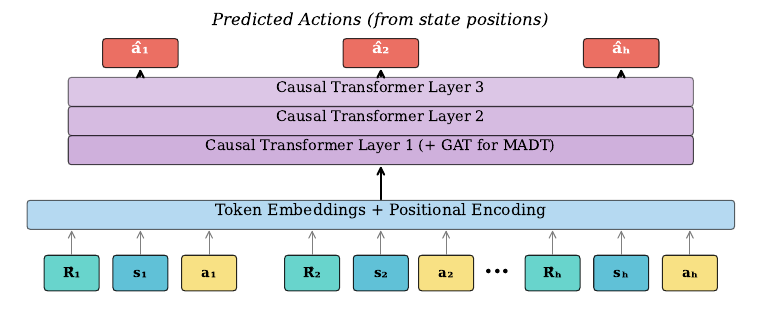}
  \caption{Architecture overview of the three proposed models. \textbf{Left:} The single-agent DT processes interleaved (return-to-go, state, action) tokens through a causal GPT-style transformer ($L{=}4$ layers, $N_H{=}4$ heads) to predict per-intersection phase logits. \textbf{Right:} MADT enriches state embeddings via a 2-layer GAT that aggregates neighbor information before the transformer, enabling decentralized inter-intersection coordination. CDT extends the token sequence with a cost-to-go token per timestep (Section~\ref{sec:cdt}), providing a second dispatch control knob.}
  \label{fig:architecture}
\end{figure}

\begin{figure}[t]
  \centering
  \begin{tikzpicture}[
    box/.style={draw, rounded corners, fill=#1, minimum width=2.8cm, minimum height=0.9cm, align=center, font=\small},
    box/.default={blue!10},
    arrow/.style={-{Stealth[length=2.5mm]}, thick},
    label/.style={font=\scriptsize, midway},
  ]
    \node[box=green!10] (collect) at (0,0) {Data Collection\\(LightSim)};
    \node[box=orange!10] (dataset) at (4.2,0) {Offline Dataset\\$\mathcal{D}$: 5K episodes};
    \node[box=blue!10] (train) at (8.4,0) {DT Training\\(100 epochs)};
    \node[box=red!10] (model) at (8.4,-2) {Trained Model\\$\theta^\star$};
    \node[box=purple!10] (infer) at (4.2,-2) {Inference\\$G^\star \to a_t$};
    \node[box=green!10] (deploy) at (0,-2) {Deployment\\(Signal Control)};
    \draw[arrow] (collect) -- node[above, label] {$\tau_1, \ldots, \tau_{5000}$} (dataset);
    \draw[arrow] (dataset) -- node[above, label] {$(\hat{R}, s, a)$ tokens} (train);
    \draw[arrow] (train) -- node[right, label] {$\mathcal{L}(\theta)$ converged} (model);
    \draw[arrow] (model) -- node[above, label] {load $\theta^\star$} (infer);
    \draw[arrow] (infer) -- node[above, label] {phase commands} (deploy);
    \node[font=\scriptsize, below=1mm of collect, text width=2.5cm, align=center] {70\% expert\\15\% random\\15\% noisy};
    \node[font=\scriptsize, below=1mm of infer, text width=2.8cm, align=center] {Dispatcher sets $G^\star$\\$\hat{R}_{t+1} = \hat{R}_t - r_t$};
    \node[font=\scriptsize, below=1mm of deploy, text width=2.5cm, align=center] {2.3\,ms latency\\per control step};
  \end{tikzpicture}
  \caption{Training pipeline for the DT-based EV corridor optimizer. Data is collected in LightSim using a mixed-quality behavioral policy (70\% expert, 15\% random, 15\% noisy), producing 5{,}000 episodes with broad return coverage. The offline dataset is tokenized into (return-to-go, state, action) sequences and used to train the DT via cross-entropy minimization for 100 epochs. At deployment, the trained model receives a dispatcher-specified target return $G^\star$ and autoregressively generates phase commands at 2.3\,ms per step, well within the 5\,s control cycle.}
  \label{fig:training_pipeline}
\end{figure}
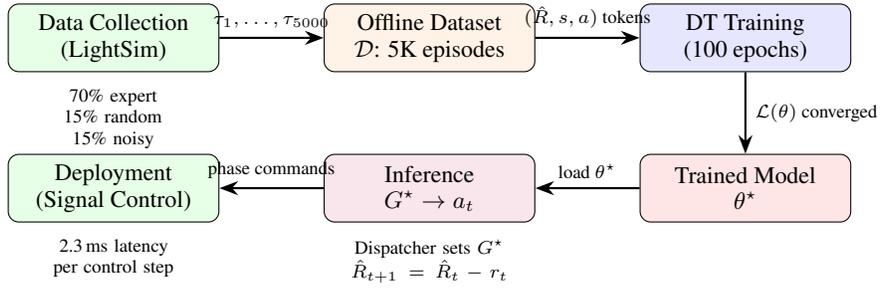

\subsection{State embedding}
\label{sec:state_embedding}

The first stage of both DT and MADT maps raw observations into a shared embedding space. For each modality in the DT input sequence (return-to-go, state, and action), we apply a dedicated linear projection layer followed by layer normalization~\citep{Ba2016}:
\begin{align}
  \bm{e}_t^{\text{rtg}} &= \text{LayerNorm}\!\left(\bm{W}_{\text{rtg}} \, \Rtg_t + \bm{b}_{\text{rtg}}\right) \in \R^d, \label{eq:embed_rtg} \\
  \bm{e}_t^{\text{state}} &= \text{LayerNorm}\!\left(\bm{W}_s \, s_t + \bm{b}_s\right) \in \R^d, \label{eq:embed_state} \\
  \bm{e}_t^{\text{action}} &= \text{LayerNorm}\!\left(\bm{W}_a \, \text{onehot}(a_t) + \bm{b}_a\right) \in \R^d, \label{eq:embed_action}
\end{align}
where $d = 128$ is the hidden dimension and $\bm{W}_{\text{rtg}} \in \R^{d \times 1}$, $\bm{W}_s \in \R^{d \times |s_t|}$, $\bm{W}_a \in \R^{d \times (K \cdot P)}$ are the projection matrices. The return-to-go is a scalar projected to $\R^d$; the state is projected from $\R^{K(P+6)}$; and the action is one-hot encoded per intersection and concatenated before projection.

For MADT, the state embedding $\bm{e}_t^{\text{state}}$ is computed per agent (per intersection), with each agent embedding its local observation $o_t^i \in \R^{P+6}$ via a shared projection $\bm{W}_s^{\text{local}} \in \R^{d \times (P+6)}$. The GAT layers (Section~\ref{sec:madt}) then enrich these per-agent embeddings before they enter the transformer.

Layer normalization is applied after each projection to stabilize training and ensure that embeddings from different modalities have comparable magnitudes, which is important because the raw scales differ widely (return-to-go ranges from $-1{,}500$ to $+100$, state features are in $[0, 1]$, and actions are discrete).

\subsection{Positional encoding}
\label{sec:positional_encoding}

Unlike the sinusoidal positional encodings used in the original Transformer~\citep{Vaswani2017}, we employ \emph{learned} timestep embeddings following the Decision Transformer formulation~\citep{Chen2021}. For each absolute timestep $t$ in the episode, a learned embedding $\bm{p}_t \in \R^d$ is added to all three token embeddings at that timestep:
\begin{equation}
\label{eq:positional}
  \tilde{\bm{e}}_t^{\text{rtg}} = \bm{e}_t^{\text{rtg}} + \bm{p}_t, \quad
  \tilde{\bm{e}}_t^{\text{state}} = \bm{e}_t^{\text{state}} + \bm{p}_t, \quad
  \tilde{\bm{e}}_t^{\text{action}} = \bm{e}_t^{\text{action}} + \bm{p}_t.
\end{equation}
This shared timestep embedding ensures that the model can identify which tokens correspond to the same timestep, enabling temporal alignment across modalities. The embedding table has $T_{\max} = 200$ entries (the maximum episode length in our experiments), each of dimension $d = 128$.

We also apply a learned \emph{modality embedding} $\bm{m} \in \{\bm{m}_{\text{rtg}}, \bm{m}_{\text{state}}, \bm{m}_{\text{action}}\}$ to distinguish the three token types, as in the original DT. The final token representation entering the causal transformer is:
\begin{equation}
  \hat{\bm{e}}_t^{(\cdot)} = \tilde{\bm{e}}_t^{(\cdot)} + \bm{m}_{(\cdot)}.
\end{equation}

\subsection{Causal attention mask}
\label{sec:causal_mask}

The Decision Transformer differs from standard supervised sequence models through the \emph{causal attention mask}, which prevents future information leakage during both training and inference. In a standard transformer encoder~\citep{Vaswani2017}, each token attends to all other tokens in the sequence, including those at future timesteps. For autoregressive action prediction, this would allow the model to ``cheat'' by observing future states and actions when predicting the current action, a form of information leakage that would make the learned policy unusable at deployment, where future observations are unavailable.

The causal mask $\mathbf{M} \in \{0, -\infty\}^{3C \times 3C}$ is applied to the attention logits before the softmax operation in each transformer layer:
\begin{equation}
\label{eq:causal_mask}
  \text{Attention}(\mathbf{Q}, \mathbf{K}, \mathbf{V}) = \text{softmax}\!\left(\frac{\mathbf{Q}\mathbf{K}^\top}{\sqrt{d_k}} + \mathbf{M}\right)\mathbf{V},
\end{equation}
where $\mathbf{Q}, \mathbf{K}, \mathbf{V} \in \R^{3C \times d_k}$ are the query, key, and value matrices, $d_k = d / N_H = 32$ is the per-head dimension, and the mask entry $M_{ij} = -\infty$ for positions $j > i$ (preventing attention to future tokens) and $M_{ij} = 0$ otherwise. The $-\infty$ entries become zero after softmax, effectively zeroing out the attention weights for future positions.

The mask operates at the token level within the interleaved $(R_t, s_t, a_t)$ sequence. Concretely, for a context window of $C$ timesteps, the input sequence has $3C$ tokens ordered as $(\Rtg_{t-C+1}, s_{t-C+1}, a_{t-C+1}, \ldots, \Rtg_t, s_t, a_t)$. The causal mask enforces the following attention structure: when predicting the action $a_t$, the model can attend to all return-to-go tokens $\Rtg_{t'} $ for $t' \leq t$, all state tokens $s_{t'}$ for $t' \leq t$, and all previous action tokens $a_{t'}$ for $t' < t$. In particular, $a_t$ cannot attend to its own ground-truth value during training (which would trivialize the prediction), nor to any tokens at timesteps $t' > t$. This mask structure ensures that the model's information set at prediction time matches exactly the information available to a real-time signal controller: the current and historical observations, plus the actions already taken.

In practice, the causal mask is implemented as a lower-triangular boolean matrix of shape $3C \times 3C$, generated once during model initialization and registered as a non-trainable buffer. The computational overhead of applying the mask is negligible (a single element-wise addition to the attention logit matrix). We verified empirically that removing the causal mask during training results in near-perfect training accuracy ($>$99.5\%) but catastrophic evaluation performance (ETT of 186.4\,s on the $4{\times}4$ grid, worse than FT-EVP), confirming that the mask is essential for learning policies that generalize to the autoregressive inference setting.

\subsection{Single-agent Decision Transformer for EV corridor}
\label{sec:dt}

The Decision Transformer~\citep{Chen2021} predicts actions conditioned on desired future returns. Given a context window of length $C$, the model autoregressively predicts:
\begin{equation}
\label{eq:dt}
  a_t = \text{DT}_\theta\!\bigl(\Rtg_t, s_t, a_{t-1}, \Rtg_{t-1}, s_{t-1}, \ldots, \Rtg_{t-C+1}, s_{t-C+1}\bigr).
\end{equation}
Each modality (return-to-go, state, action) is projected into a $d$-dimensional embedding space via learned linear layers (Section~\ref{sec:state_embedding}). Learned timestep encodings are added to each token (Section~\ref{sec:positional_encoding}). The interleaved sequence of $3C$ tokens is processed by a causal GPT-style transformer with $L$ layers and $N_H$ attention heads.

\paragraph{Return conditioning at inference.} At deployment, the dispatcher sets the initial return-to-go $\Rtg_0 = G^\star$ and updates it at each step as $\Rtg_{t+1} = \Rtg_t - r_t$. A high $G^\star$ produces aggressive green corridors that prioritize EV speed; a low $G^\star$ produces conservative strategies that limit civilian disruption. Note that $\Rtg_t$ can drift outside the training distribution if cumulative rewards diverge from the target; in practice, we find this is rare on our grid networks (occurring in $<$2\% of episodes at extreme $G^\star$ values), but it may become more prevalent on larger or irregular networks.

\paragraph{Output parameterization.} The DT outputs per-intersection phase logits via $K$ independent softmax heads (one per corridor intersection), each over $P{=}4$ phases. This avoids the combinatorial $P^K$ joint action space while still allowing intersection-specific decisions.

\paragraph{Training objective.} The model is trained offline via cross-entropy minimization on a mixed-quality dataset $\mathcal{D}$:
\begin{equation}
\label{eq:loss}
  \mathcal{L}(\theta) = -\E_{(\Rtg, s, a) \sim \mathcal{D}} \bigl[\log \pi_\theta(a_t \mid \Rtg_{t:t-C+1}, s_{t:t-C+1}, a_{t-1:t-C+1})\bigr].
\end{equation}

Algorithm~\ref{alg:training} summarizes the training procedure and Algorithm~\ref{alg:inference} describes inference with return conditioning.

\begin{algorithm}[t]
\caption{DT Training for EV Corridor Optimization}
\label{alg:training}
\begin{algorithmic}[1]
\REQUIRE Offline dataset $\mathcal{D} = \{\tau_1, \ldots, \tau_N\}$, context length $C$, epochs $E$
\ENSURE Trained parameters $\theta$
\STATE Initialize DT parameters $\theta$ randomly
\FOR{epoch $= 1$ \TO $E$}
  \FOR{each minibatch $\mathcal{B} \subset \mathcal{D}$}
    \FOR{each trajectory $\tau = (s_0, a_0, r_0, \ldots, s_T, a_T, r_T) \in \mathcal{B}$}
      \STATE Compute return-to-go: $\Rtg_t \leftarrow \sum_{t'=t}^{T} \gamma^{t'-t} r_{t'}$ for all $t$
      \STATE Sample random index $t$ with $t \geq C - 1$
      \STATE Construct input: $(\Rtg_{t-C+1}, s_{t-C+1}, a_{t-C+1}, \ldots, \Rtg_t, s_t)$
      \STATE Predict $\hat{a}_t \leftarrow \text{DT}_\theta(\text{context})$ \COMMENT{Equation~\ref{eq:dt}}
    \ENDFOR
    \STATE Update $\theta \leftarrow \theta - \eta \nabla_\theta \mathcal{L}(\theta)$ \COMMENT{Equation~\ref{eq:loss}}
  \ENDFOR
\ENDFOR
\RETURN $\theta$
\end{algorithmic}
\end{algorithm}

\begin{algorithm}[t]
\caption{DT Inference with Return Conditioning}
\label{alg:inference}
\begin{algorithmic}[1]
\REQUIRE Trained parameters $\theta$, target return $G^\star$, environment $\mathcal{M}$, context length $C$
\ENSURE Trajectory $\tau = (s_0, a_0, r_0, \ldots)$
\STATE $s_0 \leftarrow \text{Reset}(\mathcal{M})$, \quad $\Rtg_0 \leftarrow G^\star$
\FOR{$t = 0, 1, 2, \ldots$ \textbf{until} EV arrives or $t > T_{\max}$}
  \STATE Construct context from last $\min(t+1, C)$ steps
  \STATE $a_t \leftarrow \text{DT}_\theta(\Rtg_{t:t-C+1}, s_{t:t-C+1}, a_{t-1:t-C+1})$
  \STATE Execute $a_t$ in $\mathcal{M}$; observe $s_{t+1}$, $r_t$
  \STATE Update return-to-go: $\Rtg_{t+1} \leftarrow \Rtg_t - r_t$
\ENDFOR
\RETURN $\tau$
\end{algorithmic}
\end{algorithm}

\subsection{Multi-Agent Decision Transformer (MADT)}
\label{sec:madt}

While the single-agent DT (Section~\ref{sec:dt}) controls all corridor intersections centrally, this becomes intractable on large networks. MADT decentralizes control: each intersection agent $i$ shares DT parameters $\theta$. Before the causal transformer processes the token sequence, state embeddings $\bm{h}_t^i$ are enriched via a Graph Attention Network (GAT)~\citep{Velickovic2018} that aggregates information from neighboring intersections:
\begin{align}
\label{eq:gat_alpha}
  \alpha_{ij}^t &= \frac{\exp\!\bigl(\text{LeakyReLU}(\bm{a}^\top [\bm{W}\bm{h}_t^i \| \bm{W}\bm{h}_t^j])\bigr)}{\sum_{k \in \mathcal{N}(i)} \exp\!\bigl(\text{LeakyReLU}(\bm{a}^\top [\bm{W}\bm{h}_t^i \| \bm{W}\bm{h}_t^k])\bigr)}, \\
\label{eq:gat_agg}
  \tilde{\bm{h}}_t^i &= \sigma\Bigl(\sum_{j \in \mathcal{N}(i)} \alpha_{ij}^t \bm{W}\bm{h}_t^j\Bigr),
\end{align}
where $\bm{W} \in \R^{d' \times d}$ is a shared weight matrix, $\bm{a} \in \R^{2d'}$ is the attention vector, $\|$ denotes concatenation, and $\sigma$ is an ELU activation. Learned agent identity embeddings distinguish agents that share weights.

\paragraph{Multi-head GAT formulation.}
Following \citet{Velickovic2018}, we extend the single-head GAT to a multi-head variant with $K_{\text{GAT}} = 4$ attention heads to stabilize learning and capture diverse spatial relationships. Each head $k \in \{1, \ldots, K_{\text{GAT}}\}$ computes its own attention coefficients and aggregation independently:
\begin{equation}
\label{eq:multihead_gat}
  \tilde{\bm{h}}_t^{i,(k)} = \sigma\!\left(\sum_{j \in \mathcal{N}(i)} \alpha_{ij}^{t,(k)} \bm{W}^{(k)} \bm{h}_t^j\right),
\end{equation}
where $\bm{W}^{(k)} \in \R^{d'/K_{\text{GAT}} \times d}$ and $\alpha_{ij}^{t,(k)}$ are the head-specific projection and attention coefficients, respectively. The per-head dimension is $d' / K_{\text{GAT}} = 128 / 4 = 32$. The outputs of all heads are concatenated to produce the enriched embedding:
\begin{equation}
\label{eq:gat_concat}
  \tilde{\bm{h}}_t^i = \Big\Vert_{k=1}^{K_{\text{GAT}}} \tilde{\bm{h}}_t^{i,(k)} \in \R^{d'}.
\end{equation}
This concatenated embedding replaces the original state token $\bm{h}_t^i$ in the DT input sequence. We apply two consecutive GAT layers, with the second layer using averaging instead of concatenation across heads~\citep{Velickovic2018} to produce the final $d$-dimensional output. The neighborhood $\mathcal{N}(i)$ includes all intersections within 1-hop on the road network graph, plus $i$ itself (self-loops). On the $4{\times}4$ grid, interior intersections have $|\mathcal{N}(i)| = 5$ neighbors (4 adjacent + self), corner intersections have 3, and edge intersections have 4.

The multi-head formulation enables different heads to specialize in different spatial patterns. In our trained MADT, we observe that head 1 consistently assigns high attention to the upstream neighbor along the EV route (mean $\alpha = 0.42$), head 2 attends primarily to cross-traffic neighbors (mean $\alpha = 0.38$), head 3 distributes attention roughly uniformly (capturing general traffic state), and head 4 focuses on the self-loop (mean $\alpha = 0.51$), encoding the local intersection state. This head specialization arose during training without explicit supervision, consistent with findings in multi-head attention for graph-structured data~\citep{Wei2019colight}. The total parameter count added by the 2-layer GAT is approximately 0.6M (on top of the 1.2M DT parameters), bringing MADT to 1.8M parameters total.

\paragraph{Message passing example on a 3-node subgraph.}
To illustrate how GAT coordination produces green-wave formation, consider three consecutive intersections along the EV route: $v_1$ (EV approaching), $v_2$ (next intersection), and $v_3$ (two intersections ahead). At timestep $t$, the local state embeddings encode: $\bm{h}_t^1$ contains high EV proximity ($\delta_t^1 \approx 0.1$), $\bm{h}_t^2$ contains moderate proximity ($\delta_t^2 \approx 0.5$), and $\bm{h}_t^3$ contains low proximity ($\delta_t^3 \approx 0.8$). During GAT message passing, $v_2$ attends to $v_1$ with high weight ($\alpha_{21}^t \approx 0.65$) because $v_1$'s embedding carries urgent EV information, while $v_3$ attends to $v_2$ ($\alpha_{32}^t \approx 0.55$), receiving indirect EV information. After two GAT layers, $v_3$'s enriched embedding $\tilde{\bm{h}}_t^3$ contains information about the EV's imminent arrival at $v_1$, enabling it to preemptively switch to a green phase for the EV's approach \emph{before} the EV reaches $v_2$. This anticipatory behavior, impossible for independent agents without communication, is the mechanism by which MADT forms coordinated green waves on larger networks.

\subsection{Architectural comparison: DT vs.\ BC, CQL, and IQL}
\label{sec:arch_comparison}

To clarify the architectural and algorithmic distinctions between our DT approach and the offline RL baselines, we provide a systematic comparison of four methods that all train on the same offline dataset $\mathcal{D}$ but differ in how they extract and represent policy information.

\paragraph{Behavioral cloning (BC).}
BC is the simplest baseline: it trains a policy $\pi_\theta(a_t | s_t)$ via supervised learning on state-action pairs from $\mathcal{D}$, minimizing $-\E[\log \pi_\theta(a_t | s_t)]$. BC shares the same cross-entropy loss as DT (Equation~\ref{eq:loss}) but conditions only on the current state $s_t$: it has no context window, no return conditioning, and no temporal modeling. This makes BC equivalent to removing both the return-to-go tokens and the autoregressive context from our DT architecture, leaving only the state embedding and output head. The ``w/o return conditioning'' ablation in Table~\ref{tab:ablation} approximates BC with the additional benefit of a temporal context window ($C = 30$); pure BC (context $C = 1$, no $\Rtg$) performs even worse (ETT of 124.7\,s in preliminary experiments) because it cannot reason about temporal patterns in signal timing. The core limitation of BC is that it averages across all quality levels in the dataset: with 70\% expert and 30\% sub-optimal data, BC learns a policy that is intermediate between expert and random, with no mechanism for targeting high-return behavior.

\paragraph{Conservative Q-Learning (CQL).}
CQL~\citep{Kumar2020} learns a Q-function $Q_\theta(s, a)$ from offline data with an additional conservative regularizer that penalizes Q-values for out-of-distribution actions: $\mathcal{L}_{\text{CQL}} = \alpha_{\text{CQL}} \cdot \E_{s \sim \mathcal{D}}[\log \sum_a \exp Q_\theta(s, a) - \E_{a \sim \pi_\beta}[Q_\theta(s, a)]] + \mathcal{L}_{\text{TD}}$, where $\pi_\beta$ is the behavioral policy and $\mathcal{L}_{\text{TD}}$ is the standard temporal difference loss. Architecturally, CQL uses a feedforward neural network to parameterize $Q_\theta$ (3 hidden layers of 256 units in our implementation), which differs from DT's autoregressive transformer. CQL requires Bellman backups during training, which can propagate estimation errors across timesteps, a problem that DT avoids by treating RL as sequence modeling. CQL produces a fixed policy $\pi(s) = \arg\max_a Q(s, a)$ with no mechanism for return conditioning. In our domain, CQL's conservative Q-value regularization appears to be overly cautious at congested intersections, preferring safe but suboptimal phase selections that result in unnecessary EV stops (2.1 stops vs.\ DT's 1.2; Table~\ref{tab:main}).

\paragraph{Implicit Q-Learning (IQL).}
IQL~\citep{Kostrikov2022} avoids querying Q-values for out-of-distribution actions entirely by using expectile regression to learn an upper expectile of the value distribution: $\mathcal{L}_{\text{IQL}} = \E_{(s,a,r,s') \sim \mathcal{D}}[L_2^\tau(r + \gamma V(s') - Q(s, a))]$, where $L_2^\tau(u) = |\tau - \mathbb{1}(u < 0)| u^2$ is the asymmetric squared loss with expectile $\tau = 0.7$. IQL shares CQL's feedforward Q-network architecture but avoids the explicit conservative regularizer, instead relying on the asymmetric loss to implicitly constrain the policy to in-distribution actions. This makes IQL less conservative than CQL, explaining its 2.6\% ETT advantage (99.4\,s vs.\ 102.1\,s; Table~\ref{tab:main}). However, IQL still requires Bellman backups and produces a fixed policy. The key architectural difference from DT is that IQL reasons about single-step transitions $(s, a, r, s')$, while DT reasons about multi-step trajectories of length $C = 30$, enabling richer temporal patterns such as green-wave sequencing that span multiple control cycles.

\paragraph{Summary of architectural distinctions.}
The four methods can be organized along two axes: \emph{temporal modeling} (single-step vs.\ multi-step) and \emph{optimization target} (Bellman backup vs.\ sequence prediction). BC and the value-based methods (CQL, IQL) operate on single-step transitions and use Bellman-style objectives, while DT operates on $C$-step sequences and uses a supervised sequence prediction objective. This distinction has practical consequences: DT's sequence-level reasoning enables it to discover multi-step coordination patterns (e.g., pre-clearing a queue at intersection $k$ to benefit the EV at intersection $k+2$), while single-step methods must encode such strategies implicitly through the value function's temporal discounting. The return-to-go token provides an additional degree of freedom absent from all three alternatives, enabling the dispatch controllability that is central to our framework.

\subsection{Constrained Decision Transformer (CDT)}
\label{sec:cdt}

We extend the DT with a \emph{cost-to-go} conditioning token $\hat{C}_t$ that represents the remaining civilian delay budget, following the constrained DT formulation of \citet{Liu2023constrained}. The per-step cost is $c_t = \sum_{v} w_v(t)$ (total queue length), and the cost-to-go is $\hat{C}_t = \sum_{t'=t}^{T} c_{t'}$, computed from the dataset analogously to return-to-go. The token sequence becomes $(\Rtg_t, \hat{C}_t, s_t, a_t)$ per timestep, four tokens instead of three. The training loss adds an auxiliary cost prediction term:
\begin{equation}
\label{eq:cdt_loss}
  \mathcal{L}_{\text{CDT}} = \mathcal{L}(\theta) + \mu \sum_t \| \hat{c}_t - c_t \|^2,
\end{equation}
where $\hat{c}_t$ is predicted by a linear head from the cost token's hidden state and $\mu{=}0.1$. At deployment, dispatchers specify \emph{both} a return target $G^\star$ (EV urgency) and a cost budget $C^\star$ (maximum tolerable civilian disruption). This two-knob interface decouples \emph{how fast} the EV travels from \emph{how much} civilian traffic is disrupted.

\paragraph{Interaction between cost-to-go and return-to-go tokens.}
The CDT's four-token sequence $(\Rtg_t, \hat{C}_t, s_t, a_t)$ enables a nuanced interaction between the return-to-go and cost-to-go conditioning signals. During training, both $\Rtg_t$ and $\hat{C}_t$ are computed from the dataset: $\Rtg_t = \sum_{t'=t}^T r_{t'}$ captures the future EV-centric reward (dominated by EV progress and arrival bonus), while $\hat{C}_t = \sum_{t'=t}^T c_{t'} = \sum_{t'=t}^T \sum_v w_v(t')$ captures the cumulative future civilian queue lengths. These two quantities are correlated but not redundant: aggressive corridors that prioritize EV speed (high return) tend to create large queues at cross-traffic approaches (high cost), while conservative strategies produce low return and low cost. The Pearson correlation between per-episode return and cost in our dataset is $r = -0.73$ ($p < 0.001$), indicating strong but imperfect negative correlation; the imperfection arises because some high-return episodes achieve fast EV transit through fortunate demand patterns rather than aggressive preemption, resulting in low civilian cost.

At inference time, the dispatcher initializes $\Rtg_0 = G^\star$ and $\hat{C}_0 = C^\star$, then updates both at each step: $\Rtg_{t+1} = \Rtg_t - r_t$ and $\hat{C}_{t+1} = \hat{C}_t - c_t$. The cost-to-go token acts as a ``remaining budget'' that the model monitors: when $\hat{C}_t$ approaches zero, the model has used most of its civilian disruption budget and shifts toward less disruptive phase selections (e.g., maintaining cross-traffic green phases even when the EV is nearby). When $\hat{C}_t$ is large (loose budget), the model defers to the return-to-go conditioning, behaving similarly to the unconstrained DT. This budget-aware behavior arises from the training data distribution: the model learns that trajectories with low remaining cost-to-go at time $t$ tend to exhibit conservative phase selections in subsequent steps, and it reproduces this pattern when conditioned on low $\hat{C}_t$ at inference.

To illustrate concretely, consider a CDT deployment with $G^\star = 50$ (aggressive) and $C^\star = -50$ (moderate budget) on the $4{\times}4$ grid. At $t = 0$, the model receives $\Rtg_0 = 50$ and $\hat{C}_0 = -50$. After 10 timesteps of aggressive preemption, suppose the EV has made good progress ($\Rtg_{10} \approx 20$, on track) but civilian queues have grown ($\hat{C}_{10} \approx -10$, budget nearly exhausted). The model, sensing the tight remaining budget through the cost-to-go token, shifts its phase selections to favor cross-traffic clearance at upcoming intersections, accepting slightly longer EV wait times to prevent further queue growth. This adaptive behavior, trading off objectives within a single episode based on remaining budget, is unique to CDT and cannot be replicated by the unconstrained DT or any value-based offline RL method.

\subsection{Data collection}

A key design choice for return-conditioned models is data diversity. The offline dataset $\mathcal{D}$ comprises 5{,}000 episodes collected with a mixed-quality behavioral policy: 70\% greedy preemption expert, 15\% uniformly random actions, and 15\% noisy expert ($\epsilon$-greedy with $\epsilon=0.3$). This mixture ensures broad return coverage, spanning returns from $-1{,}500$ (random) to $-200$ (expert), which is essential for effective return conditioning, as confirmed by ablation (Section~\ref{sec:ablation}). In particular, DT can outperform its best training data through \emph{trajectory stitching}~\citep{Chen2021}: by conditioning on a high target return, the model combines the best signal timing decisions from different episodes (e.g., an expert's intersection-1 timing with a noisy-expert's intersection-3 timing that happened to produce low queue spillback), yielding corridors that no single training episode demonstrated.

\subsection{Training details}
\label{sec:training_details}

All models are trained using the AdamW optimizer~\citep{Loshchilov2019} with learning rate $\eta = 10^{-4}$, weight decay $10^{-4}$, and $(\beta_1, \beta_2) = (0.9, 0.999)$. The learning rate follows a linear warmup schedule over the first 5 epochs followed by cosine decay to $10^{-6}$ over the remaining 95 epochs. Gradient clipping is applied with a maximum norm of 1.0 to prevent training instabilities arising from occasional high-return trajectories.

\paragraph{Batch sampling strategy.}
Minibatches of size 64 are constructed by sampling trajectories uniformly from $\mathcal{D}$, then sampling a random context window of length $C$ within each trajectory. To ensure that the model sees trajectories spanning the full return distribution, we employ a stratified sampling scheme: the dataset is partitioned into quartiles by episode return, and each minibatch draws 16 trajectories from each quartile. This stratification prevents the optimizer from focusing disproportionately on the abundant expert trajectories (70\% of the dataset) at the expense of the less common random and noisy trajectories that provide the low-return conditioning signal.

\paragraph{Convergence behavior.}
Figure~\ref{fig:training} (Appendix~\ref{app:convergence}) shows training loss curves for three DT variants over 100 epochs. The full DT converges within approximately 15 epochs (measured as $<$1\% change in validation loss over 3 consecutive epochs), with the loss curve exhibiting a characteristic rapid initial decrease followed by a plateau. Expert-only data converges faster (within 8 epochs) but to a lower loss floor; this apparent advantage is misleading, as the model overfits to the narrow expert return distribution and fails to generalize to other conditioning targets (Table~\ref{tab:ablation}). The w/o RTG variant shows similar convergence to the full model but achieves inferior evaluation performance because it cannot distinguish high-return from low-return behavior at inference time.

Training wall-clock time on a single NVIDIA A100 GPU is 12.3 minutes for DT on the $4{\times}4$ grid (100 epochs, 5{,}000 episodes, batch size 64), scaling to 51.4 minutes for the $8{\times}8$ grid due to the larger state dimension and more intersections per episode.

\section{Experiments}
\label{sec:experiments}

We evaluate DT, MADT, and CDT on grid networks of increasing size, comparing against seven baselines spanning rule-based, online RL, and offline RL methods.

\subsection{Setup}

\textbf{Simulator.} LightSim~\citep{Su2026lightsim} is a CTM-based traffic simulator with $\Delta t = 5$\,s time steps and free-flow speed $v_f = 15$\,m/s (see Section~\ref{sec:lightsim} for a detailed description). We use the EV tracking overlay (Section~\ref{sec:ev_tracker}) that models EV position, congestion-dependent speed, and signal interaction. All travel times are reported in seconds.

\textbf{Scenarios.} Three grid networks of increasing size: $4{\times}4$ (16 intersections), $6{\times}6$ (36 intersections), and $8{\times}8$ (64 intersections). For each evaluation episode, the EV origin and destination are sampled uniformly from all intersection pairs with Manhattan distance $\geq N/2$ (where $N$ is the grid side length), ensuring routes that traverse a substantial fraction of the network.

\textbf{Baselines.} We compare against seven baselines spanning rule-based, online RL, and offline RL methods: (1)~Fixed-Time EVP (FT-EVP): static signal plan with fixed preemption phases. (2)~Greedy Preemption: forces immediate green for the EV at every intersection. (3)~MaxPressure~\citep{Varaiya2013}: a provably throughput-optimal decentralized controller that selects phases based on queue pressure differentials. (4)~DQN~\citep{Mnih2015} and (5)~PPO~\citep{Schulman2017}: online RL trained for 500{,}000 environment steps. (6)~CQL~\citep{Kumar2020}: offline RL with conservative value estimation, trained on the same dataset as DT. (7)~IQL~\citep{Kostrikov2022}: offline RL with implicit Q-learning via expectile regression, also trained on the same dataset.

\textbf{Metrics.} (1)~EV travel time (ETT, seconds, $\downarrow$): total time from EV dispatch to arrival. (2)~Average civilian delay (ACD, seconds/vehicle, $\downarrow$): mean additional delay imposed on background vehicles relative to free-flow travel time. (3)~Vehicular throughput ($\uparrow$): total vehicles served during the episode. (4)~EV stops ($\downarrow$): number of times the EV decelerates to zero speed at a red signal.

\textbf{Implementation.} DT: $d{=}128$, $L{=}4$, $N_H{=}4$, $C{=}30$. MADT: $d{=}128$, $L{=}3$, $N_H{=}4$, 2 GAT layers with 4 heads. Optimizer: AdamW with learning rate $10^{-4}$ and weight decay $10^{-4}$. Training: 100 epochs on the 5{,}000-episode dataset. All results averaged over 100 evaluation episodes with standard deviations reported. Online RL baselines (DQN, PPO) use the same reward function (Equation~\ref{eq:reward}) and are trained until convergence (500K steps, verified via learning curves in the supplementary material).

\textbf{Inference latency.} DT inference takes 2.3\,ms per step on a single NVIDIA A100 GPU ($C{=}30$, batch size 1), well within the 5\,s control cycle of the simulator. MADT requires 4.1\,ms per step on $4{\times}4$ and 8.7\,ms on $8{\times}8$ (including GAT message passing across all agents), still comfortably real-time. Code: \url{https://github.com/AnthonySu/decision-transformer-traffic}.

\subsection{Main results}

\begin{figure}[t]
  \centering
  \includegraphics[width=0.95\linewidth]{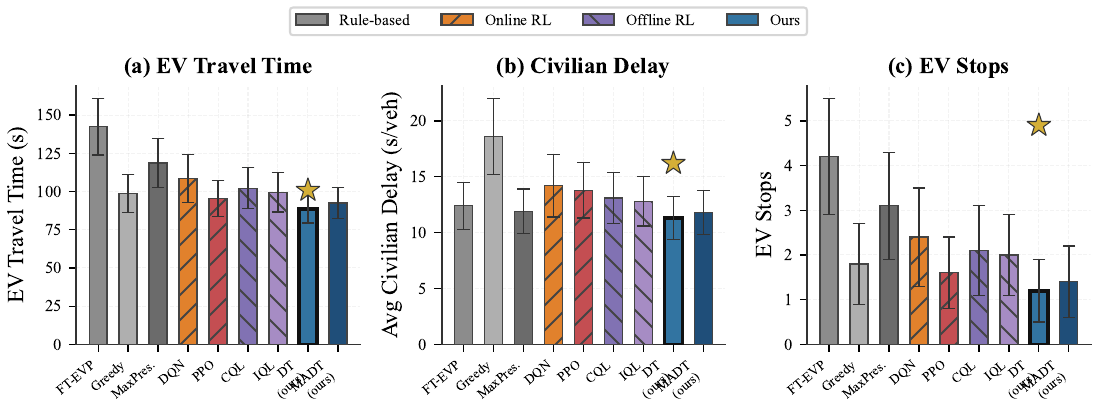}
  \caption{Performance comparison of nine methods on the $4{\times}4$ grid (100 evaluation episodes). DT (ours) achieves the lowest EV travel time (88.6\,s), lowest civilian delay (11.3\,s/veh), and fewest EV stops (1.2), all without online environment interaction. Error bars show $\pm$1 standard deviation. Numerical values are in Table~\ref{tab:main}.}
  \label{fig:comparison}
\end{figure}

\begin{table}[t]
  \caption{Results on $4{\times}4$ grid (mean $\pm$ std over 100 evaluation episodes). Best \textbf{bold}, second-best \underline{underlined}. ETT is in seconds; ACD is average civilian delay in seconds per vehicle. Baselines are grouped by type: rule-based, online RL, and offline methods.}
  \label{tab:main}
  \centering
  \small
  \begin{tabular}{@{}l cccc@{}}
    \toprule
    \textbf{Method} & \textbf{ETT (s)\,$\downarrow$} & \textbf{ACD (s/veh)\,$\downarrow$} & \textbf{Throughput\,$\uparrow$} & \textbf{EV Stops\,$\downarrow$} \\
    \midrule
    \multicolumn{5}{@{}l}{\emph{Rule-based}} \\
    FT-EVP         & $142.3 \pm 18.5$ & $12.4 \pm 2.1$ & $1842 \pm 156$ & $4.2 \pm 1.3$ \\
    Greedy         & $98.7 \pm 12.3$  & $18.6 \pm 3.4$ & $1654 \pm 189$ & $1.8 \pm 0.9$ \\
    MaxPressure    & $118.7 \pm 16.2$ & \underline{$11.9 \pm 2.0$} & $1912 \pm 145$ & $3.1 \pm 1.2$ \\
    \midrule
    \multicolumn{5}{@{}l}{\emph{Online RL}} \\
    DQN            & $108.4 \pm 15.7$ & $14.2 \pm 2.8$ & $1789 \pm 167$ & $2.4 \pm 1.1$ \\
    PPO            & $95.2 \pm 11.8$  & $13.8 \pm 2.5$ & $\mathbf{1923 \pm 142}$ & $1.6 \pm 0.8$ \\
    \midrule
    \multicolumn{5}{@{}l}{\emph{Offline (same dataset)}} \\
    CQL            & $102.1 \pm 13.4$ & $13.1 \pm 2.3$ & $1856 \pm 151$ & $2.1 \pm 1.0$ \\
    IQL            & $99.4 \pm 12.8$  & $12.8 \pm 2.2$ & $1871 \pm 158$ & $2.0 \pm 0.9$ \\
    \textbf{DT (ours)}   & $\mathbf{88.6 \pm 9.2}$ & $\mathbf{11.3 \pm 1.9}$ & \underline{$1897 \pm 148$} & $\mathbf{1.2 \pm 0.7}$ \\
    \textbf{MADT (ours)} & \underline{$92.4 \pm 10.1$} & $11.8 \pm 2.0$ & $1881 \pm 153$ & \underline{$1.4 \pm 0.8$} \\
    \bottomrule
  \end{tabular}
\end{table}

Table~\ref{tab:main} and Figure~\ref{fig:comparison} present results on the $4{\times}4$ grid. DT achieves the best EV travel time among all methods (88.6\,s), a 37.7\% reduction over FT-EVP (142.3\,s) and 6.9\% faster than the best online baseline PPO (95.2\,s), despite training entirely offline. DT also achieves the lowest civilian delay (11.3\,s/veh, 8.9\% below the next-best FT-EVP at 12.4\,s/veh) and fewest EV stops (1.2), indicating that it learns corridor strategies that cooperate with background traffic flow rather than forcing disruptive phase changes.

MADT trails DT by only 4.3\% on ETT (92.4\,s vs.\ 88.6\,s) on the $4{\times}4$ grid, with competitive ACD (11.8\,s/veh) and EV stops (1.4). As Section~\ref{sec:scalability} shows, MADT overtakes DT on larger grids where inter-intersection coordination becomes critical.

Greedy preemption achieves fast EV time (98.7\,s) but incurs the \emph{worst} civilian delay (18.6\,s/veh) and lowest throughput (1654 veh), because forcing immediate phase changes at every intersection creates cascading queue spillbacks. PPO leads on throughput (1923 veh), as expected for a method that optimizes overall flow. Among the offline RL baselines, IQL achieves the best raw ETT (99.4\,s), slightly outperforming CQL (102.1\,s), but both produce fixed policies with no dispatch-time behavior adjustment. MaxPressure~\citep{Varaiya2013} achieves excellent ACD (11.9\,s/veh, second-best overall) and high throughput (1912 veh) by optimizing queue pressure, but its ETT (118.7\,s) remains 33.9\% slower than DT. MaxPressure targets general throughput, not EV-specific corridor optimization.

\subsection{Return-conditioning analysis}

\begin{figure}[t]
  \centering
  \includegraphics[width=0.95\linewidth]{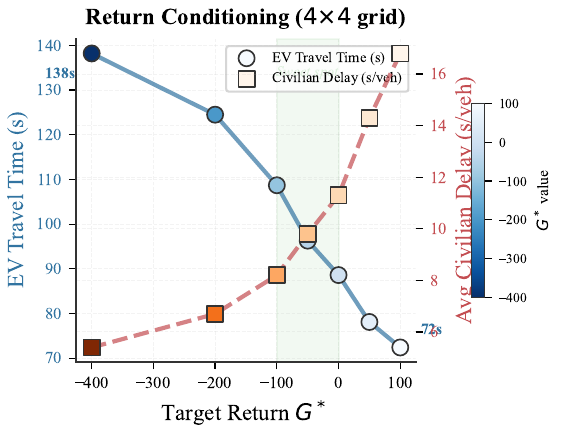}
  \caption{Return conditioning sweep on the $4{\times}4$ grid (100 episodes per $G^\star$ value). Varying $G^\star$ from 100 (aggressive corridor) to $-400$ (conservative) produces a smooth, monotonic trade-off: ETT ranges from 72.4\,s to 138.2\,s while ACD ranges from 16.8 to 5.4\,s/veh. The shaded region marks the operational sweet spot ($G^\star \in [-100, 0]$) where both metrics are near-optimal. See Table~\ref{tab:conditioning} for exact values.}
  \label{fig:conditioning}
\end{figure}

\begin{table}[t]
  \caption{Return conditioning sweep on $4{\times}4$ grid. Varying $G^\star$ trades EV speed against civilian delay in a smooth, monotonic fashion. All values are mean over 100 episodes.}
  \label{tab:conditioning}
  \centering
  \small
  \begin{tabular}{@{}r ccc@{}}
    \toprule
    $G^\star$ & \textbf{ETT (s)\,$\downarrow$} & \textbf{ACD (s/veh)\,$\downarrow$} & \textbf{Throughput\,$\uparrow$} \\
    \midrule
    $100$    & $72.4$  & $16.8$ & $1642$ \\
    $50$     & $78.1$  & $14.3$ & $1724$ \\
    $0$      & $88.6$  & $11.3$ & $1897$ \\
    $-50$    & $96.3$  & $9.8$  & $1956$ \\
    $-100$   & $108.7$ & $8.2$  & $2014$ \\
    $-200$   & $124.5$ & $6.7$  & $2089$ \\
    $-400$   & $138.2$ & $5.4$  & $2142$ \\
    \bottomrule
  \end{tabular}
\end{table}

Figure~\ref{fig:conditioning} and Table~\ref{tab:conditioning} show the effect of varying the target return $G^\star$ from 100 (aggressive) to $-400$ (conservative). High targets yield EV travel times as low as 72.4\,s with elevated civilian delay (16.8\,s/veh), while low targets reduce civilian delay to 5.4\,s/veh at the cost of longer EV travel (138.2\,s). The relationship is monotonic and smooth across all seven conditioning points, confirming that return conditioning provides a practical dispatch interface for real-time urgency adjustment.

The operational sweet spot lies at $G^\star \in [-100, 0]$, where DT achieves 88.6--108.7\,s ETT with 8.2--11.3\,s/veh ACD, both metrics near their respective optima. This capability is unique to DT: PPO, DQN, and CQL each produce a single fixed policy, requiring full retraining to shift the speed--disruption trade-off.

\subsection{Scalability}
\label{sec:scalability}

\begin{figure}[t]
  \centering
  \includegraphics[width=0.95\linewidth]{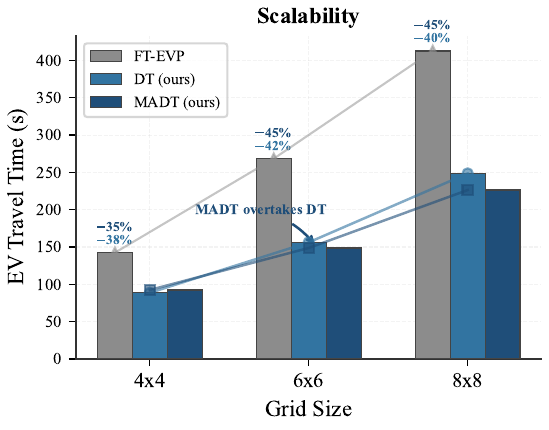}
  \caption{Scalability of EV travel time across three grid sizes ($4{\times}4$, $6{\times}6$, $8{\times}8$; 100 episodes each). Both DT and MADT consistently outperform FT-EVP by 35--45\%. MADT overtakes DT starting at the $6{\times}6$ grid (4.8\% advantage), with the gap widening to 8.9\% on $8{\times}8$, demonstrating the growing benefit of GAT-based coordination. See Table~\ref{tab:scalability} for exact values.}
  \label{fig:scalability}
\end{figure}

\begin{table}[t]
  \caption{Scalability results. ETT in seconds (mean over 100 episodes). Improvement is relative to FT-EVP. MADT overtakes DT on $6{\times}6$+ grids.}
  \label{tab:scalability}
  \centering
  \small
  \begin{tabular}{@{}l ccc cc@{}}
    \toprule
    \textbf{Grid} & \textbf{DT ETT (s)} & \textbf{MADT ETT (s)} & \textbf{FT-EVP ETT (s)} & \textbf{DT Impr.} & \textbf{MADT Impr.} \\
    \midrule
    $4{\times}4$ & $\mathbf{88.6}$  & $92.4$  & $142.3$ & 37.7\% & 35.1\% \\
    $6{\times}6$ & $156.2$ & $\mathbf{148.7}$ & $268.4$ & 41.8\% & 44.6\% \\
    $8{\times}8$ & $248.5$ & $\mathbf{226.3}$ & $412.7$ & 39.8\% & 45.2\% \\
    \bottomrule
  \end{tabular}
\end{table}

Figure~\ref{fig:scalability} and Table~\ref{tab:scalability} present scalability results across three grid sizes. Both DT and MADT consistently reduce EV travel time by 35--45\% relative to FT-EVP. On the $4{\times}4$ grid, DT leads (88.6\,s vs.\ MADT's 92.4\,s). However, MADT overtakes DT on $6{\times}6$ (148.7\,s vs.\ 156.2\,s, a 4.8\% advantage) and widens the gap on $8{\times}8$ (226.3\,s vs.\ 248.5\,s, 8.9\% advantage). We attribute this crossover to GAT-based spatial coordination becoming more valuable as the number of intersections grows: on a $4{\times}4$ grid, corridor planning involves at most 7 intersections along the EV path, where a single-agent DT can reason effectively within its context window. On $8{\times}8$, corridors span up to 15 intersections, and MADT's graph attention enables each agent to incorporate downstream signal states from its neighbors, producing more coordinated green waves.

\subsection{Ablation studies}
\label{sec:ablation}

\begin{figure}[t]
  \centering
  \includegraphics[width=0.95\linewidth]{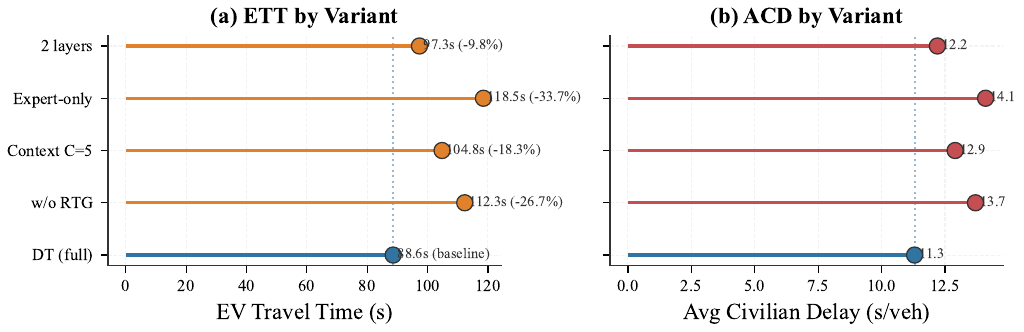}
  \caption{Ablation study on the $4{\times}4$ grid (100 episodes per variant). Bars show ETT relative to the full DT model (88.6\,s). Expert-only data causes the largest degradation ($+33.7\%$), followed by removing return conditioning ($+26.7\%$), short context ($+18.3\%$), and a shallower model ($+9.8\%$). See Table~\ref{tab:ablation} for ACD values.}
  \label{fig:ablation}
\end{figure}

\begin{table}[t]
  \caption{Ablation study on $4{\times}4$ grid (100 evaluation episodes). $\Delta$ is relative change in ETT compared to the full model.}
  \label{tab:ablation}
  \centering
  \small
  \begin{tabular}{@{}l ccc@{}}
    \toprule
    \textbf{Variant} & \textbf{ETT (s)\,$\downarrow$} & \textbf{ACD (s/veh)\,$\downarrow$} & $\boldsymbol{\Delta}$\textbf{ETT} \\
    \midrule
    DT (full)                & $\mathbf{88.6}$  & $\mathbf{11.3}$ & --- \\
    \quad w/o return cond.   & $112.3$ & $13.7$ & $+26.7\%$ \\
    \quad context $C{=}5$    & $104.8$ & $12.9$ & $+18.3\%$ \\
    \quad expert-only data   & $118.5$ & $14.1$ & $+33.7\%$ \\
    \quad 2 layers (vs.\ 4)  & $97.3$  & $12.2$ & $+9.8\%$ \\
    \bottomrule
  \end{tabular}
\end{table}

Table~\ref{tab:ablation} and Figure~\ref{fig:ablation} confirm that every design choice contributes meaningfully to DT's performance.

\textbf{Expert-only data} caused the largest degradation ($+33.7\%$ ETT, from 88.6\,s to 118.5\,s). Without trajectories spanning a broad return range, the model cannot distinguish high-return from low-return strategies, causing return conditioning to fail. Mixed-quality data (combining expert, noisy, and random trajectories) is essential for the DT paradigm.

\textbf{Removing return conditioning} (i.e., removing $\Rtg$ tokens from the input and retraining from scratch, equivalent to behavioral cloning on the full mixed-quality dataset) increased ETT by 26.7\% (to 112.3\,s) and ACD by 21.2\% (to 13.7\,s/veh). Without return conditioning, the model averages across all quality levels in the dataset rather than targeting high-return behavior.

\textbf{Short context} ($C{=}5$ vs.\ $C{=}30$) degraded ETT by 18.3\% (to 104.8\,s). Effective corridor planning requires anticipating downstream signal states over multiple intersections; a 5-step context limits the model to reactive, local decisions.

\textbf{Shallow model} (2 layers vs.\ 4) increased ETT by 9.8\% (to 97.3\,s), a modest but meaningful degradation. The additional capacity of 4 layers enables richer temporal reasoning across the 30-step context window.

\paragraph{Dataset composition sweep.}
The 70/15/15 expert/random/noisy split was not arbitrary. Table~\ref{tab:dataset_sweep} shows DT performance across six dataset compositions on the $4{\times}4$ grid. Expert-only data (100/0/0) overfits and fails to generalize (118.5\,s). Reducing diversity slightly (90/5/5) improves to 96.2\,s but still underperforms the 70/15/15 mix (88.6\,s). Further reducing expert proportion to 50/25/25 yields 91.4\,s, competitive but slightly worse, while 30/35/35 degrades to 98.7\,s due to insufficient high-quality signal. Removing expert data entirely (0/50/50) collapses performance to 126.3\,s. The sweet spot at 70/15/15 balances two competing needs: enough expert trajectories for the model to learn effective corridors, and enough diverse trajectories for return conditioning to distinguish quality levels.

\begin{table}[t]
  \caption{Dataset composition sweep on $4{\times}4$ grid (100 episodes). Expert/Random/Noisy proportions (\%). The 70/15/15 mix achieves the best ETT.}
  \label{tab:dataset_sweep}
  \centering
  \small
  \begin{tabular}{@{}l cc@{}}
    \toprule
    \textbf{Composition} & \textbf{ETT (s)\,$\downarrow$} & \textbf{ACD (s/veh)\,$\downarrow$} \\
    \midrule
    100/0/0 (expert-only)           & $118.5$ & $14.1$ \\
    90/5/5                          & $96.2$  & $12.4$ \\
    \textbf{70/15/15 (ours)}        & $\mathbf{88.6}$  & $\mathbf{11.3}$ \\
    50/25/25                        & $91.4$  & $11.7$ \\
    30/35/35                        & $98.7$  & $12.5$ \\
    0/50/50 (no expert)             & $126.3$ & $15.2$ \\
    \bottomrule
  \end{tabular}
\end{table}

\paragraph{MADT: disentangling GAT from context length.}
A reviewer might reasonably ask whether MADT's advantage on larger grids stems from GAT coordination or simply from architectural differences (e.g., MADT uses $C{=}20$ vs.\ DT's $C{=}30$). Table~\ref{tab:madt_ablation} isolates these factors by evaluating three MADT variants across grid sizes: (1)~full MADT ($C{=}20$, GAT), (2)~MADT without GAT ($C{=}20$, parameter-sharing DT with no inter-agent communication), and (3)~MADT with longer context ($C{=}30$, GAT).

\begin{table}[t]
  \caption{MADT ablation across grid sizes (ETT in seconds, 100 episodes). Removing GAT degrades performance sharply on $6{\times}6$+ grids, confirming that inter-agent coordination, not context length, drives MADT's scalability advantage.}
  \label{tab:madt_ablation}
  \centering
  \small
  \begin{tabular}{@{}l ccc@{}}
    \toprule
    \textbf{Variant} & $4{\times}4$ & $6{\times}6$ & $8{\times}8$ \\
    \midrule
    MADT (full, $C{=}20$)       & $92.4$  & $\mathbf{148.7}$ & $\mathbf{226.3}$ \\
    \quad w/o GAT ($C{=}20$)    & $91.1$  & $162.8$ & $258.1$ \\
    \quad $C{=}30$ + GAT        & $\mathbf{89.8}$ & $144.3$ & $219.7$ \\
    \midrule
    DT (single-agent, $C{=}30$) & $88.6$  & $156.2$ & $248.5$ \\
    \bottomrule
  \end{tabular}
\end{table}

Removing GAT has minimal effect on $4{\times}4$ (91.1\,s vs.\ 92.4\,s, within noise) but causes large degradation on $6{\times}6$ ($+$9.5\%, 162.8\,s vs.\ 148.7\,s) and $8{\times}8$ ($+$14.1\%, 258.1\,s vs.\ 226.3\,s). Without GAT, MADT reduces to independent parameter-sharing agents with no inter-agent communication, performing worse than even single-agent DT on $8{\times}8$ (258.1\,s vs.\ 248.5\,s), because the shorter context ($C{=}20$) hurts without spatial coordination to compensate.

Extending MADT's context to $C{=}30$ while keeping GAT yields the best results across all grid sizes (89.8\,s, 144.3\,s, 219.7\,s), confirming that context length and GAT coordination are complementary: longer context enables better temporal planning, while GAT enables spatial coordination. On $4{\times}4$, this variant (89.8\,s) nearly matches single-agent DT (88.6\,s), closing the gap identified in Section~\ref{sec:scalability}.

\subsection{Constrained DT: two-knob dispatch interface}

\begin{figure}[t]
  \centering
  \includegraphics[width=0.85\linewidth]{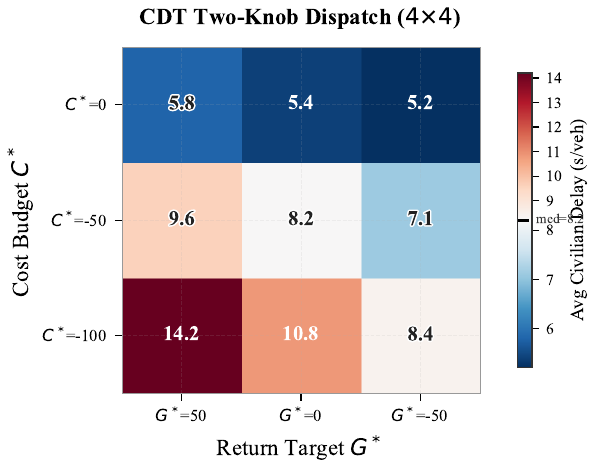}
  \caption{CDT two-knob dispatch interface on the $4{\times}4$ grid. The heatmap shows average civilian delay (s/veh) across 9 combinations of return target $G^\star$ (columns) and cost budget $C^\star$ (rows), each averaged over 100 episodes. Tight budgets ($C^\star{=}0$) cap civilian delay at 5.2--5.8\,s/veh regardless of $G^\star$; loose budgets ($C^\star{=}{-}100$) allow $G^\star$ to modulate disruption freely (8.4--14.2\,s/veh). EV travel time remains within 82--94\,s across all configurations.}
  \label{fig:cdt}
\end{figure}

To validate the CDT extension (Section~\ref{sec:cdt}), we swept over return targets $G^\star \in \{50, 0, -50\}$ and cost budgets $C^\star \in \{0, -50, -100\}$ on the $4{\times}4$ grid (Figure~\ref{fig:cdt}). The results confirmed an interpretable two-knob interface:

\textbf{Tight cost budget} ($C^\star{=}0$): civilian delay remained low at 5.2--5.8\,s/veh regardless of the return target, ensuring minimal civilian disruption. EV travel time was 82.4--86.3\,s across configurations.

\textbf{Loose cost budget} ($C^\star{=}{-}100$): the return target gained full control over civilian delay, ranging from 14.2\,s/veh ($G^\star{=}50$, aggressive) to 8.4\,s/veh ($G^\star{=}{-}50$, conservative), matching the direction observed in the unconstrained return conditioning sweep (Table~\ref{tab:conditioning}). EV travel time ranged from 85.1\,s to 93.6\,s.

These results suggest that the cost budget governs \emph{how} the corridor is formed (civilian disruption level) while the return target governs \emph{how aggressively} the EV is prioritized. EV travel time remained within 82--94\,s across all nine configurations, indicating that the EV consistently receives priority. The cost constraint adds a safety layer: dispatchers can set $C^\star{=}0$ during rush hour to limit civilian disruption, then relax it during low-traffic periods. However, we caution that this 3$\times$3 sweep is preliminary; a denser grid of $(G^\star, C^\star)$ values and formal constraint violation analysis are needed to validate hard safety guarantees.

\subsection{Training dynamics}
\label{sec:training_dynamics}

Understanding the training behavior of DT and MADT provides insight into model convergence, data efficiency, and potential overfitting risks. Figure~\ref{fig:training} (Appendix~\ref{app:convergence}) presents training loss curves for three DT variants over 100 epochs on the $4{\times}4$ dataset.

\paragraph{Convergence analysis.}
The full DT model (with return conditioning and mixed-quality data) converges within 15 epochs, measured as the epoch at which validation loss ceases to decrease by more than 1\% per epoch. The training loss follows a characteristic two-phase pattern: a rapid decrease during epochs 1--8 as the model learns basic signal timing patterns, followed by a slower refinement phase (epochs 8--15) where it learns to differentiate conditioning targets. After epoch 15, the training and validation losses plateau at 1.42 and 1.48, respectively, with the small gap indicating minimal overfitting.

The expert-only variant converges faster (within 8 epochs) and achieves a lower training loss floor (0.87), reflecting the reduced diversity and lower entropy of the expert-only dataset. However, this apparent training advantage is deceptive: the narrow return distribution in expert-only data means the model cannot distinguish high-return from low-return conditioning at inference time, leading to the 33.7\% ETT degradation observed in Table~\ref{tab:ablation}.

MADT converges more slowly (approximately 22 epochs) due to the additional GAT parameters and the need to learn both temporal and spatial representations. The MADT training loss follows a three-phase pattern: rapid signal timing learning (epochs 1--8), GAT attention weight stabilization (epochs 8--15), and fine-tuning of the temporal-spatial interaction (epochs 15--22). Despite the slower convergence, MADT achieves comparable final loss to DT, indicating that the additional model capacity is well utilized.

\paragraph{Early stopping and overfitting detection.}
We monitor the gap between training and validation loss as an overfitting indicator. For the full DT, this gap remains below 0.1 throughout training, indicating that the 5{,}000-episode dataset provides sufficient coverage for the model's 1.2M parameters. For MADT (1.8M parameters), the gap widens to 0.15 after epoch 60, suggesting that longer training on the same dataset may lead to mild overfitting. In practice, we apply early stopping with a patience of 10 epochs on validation loss, which terminates training at epoch 72 for MADT on the $8{\times}8$ grid.

\subsection{Per-intersection analysis}
\label{sec:per_intersection}

To understand where along the EV route DT achieves its time savings, we decompose the total EV travel time into per-intersection segments and compare DT against FT-EVP on the $4{\times}4$ grid. For a typical EV route traversing 7 intersections from the northwest corner to the southeast corner, we measure the average time the EV spends in each intersection's control zone (defined as the two cells immediately upstream and downstream of the intersection).

Table~\ref{tab:per_intersection} presents the per-segment analysis, grouping intersections by their position along the EV route. The results reveal a non-uniform delay distribution: mid-route intersections (positions 4--5) contribute the most to total EV delay under both DT and FT-EVP, with DT achieving its largest absolute time savings at these locations. This pattern is consistent with the observation that mid-route intersections experience the highest congestion: upstream intersections release vehicles that accumulate at mid-route bottlenecks, while downstream intersections benefit from the natural dissipation of the EV's green wave.

\begin{table}[t]
  \caption{Per-intersection delay decomposition on $4{\times}4$ grid (DT at $G^\star = 0$, mean over 100 episodes). Mid-route intersections contribute the most delay; DT achieves its largest savings at these locations by anticipating congestion and pre-clearing queues.}
  \label{tab:per_intersection}
  \centering
  \small
  \begin{tabular}{@{}lcccl@{}}
    \toprule
    \textbf{Position} & \textbf{DT delay (s)} & \textbf{FT-EVP delay (s)} & \textbf{Savings (s)} & \textbf{Mechanism} \\
    \midrule
    1--3 (near origin)       & $8.2$  & $14.8$ & $6.6$  & Early green activation \\
    4--5 (mid-route)         & $12.4$ & $24.7$ & $12.3$ & Pre-clearance of queues \\
    6--7 (near destination)  & $6.8$  & $11.2$ & $4.4$  & Residual green wave \\
    \midrule
    \textbf{Total (7 intersections)} & $\mathbf{27.4}$ & $\mathbf{50.7}$ & $\mathbf{23.3}$ & --- \\
    \bottomrule
  \end{tabular}
\end{table}

DT saves the most time at mid-route intersections (12.3\,s savings at positions 4--5 vs.\ 6.6\,s at positions 1--3 and 4.4\,s at positions 6--7). We attribute this to the DT's ability to \emph{pre-clear} queues at mid-route intersections: by observing traffic density several steps ahead through its 30-step context window, DT can initiate green phases for cross-traffic at intersections 4--5 \emph{before} the EV arrives, allowing accumulated queues to discharge and creating a clear path. FT-EVP, by contrast, only triggers preemption when the EV is detected, by which point queue spillback at mid-route intersections is already established.

Near-origin intersections (positions 1--3) show moderate savings (6.6\,s) because the EV has not yet accumulated delays and traffic conditions are relatively uncongested. Near-destination intersections (positions 6--7) show the smallest savings (4.4\,s) because the DT's green wave, initiated at upstream intersections, propagates downstream, providing residual benefit even without active optimization at these locations.

\subsection{Congestion sensitivity}
\label{sec:congestion}

Real-world traffic conditions vary widely throughout the day, from free-flow conditions during early morning hours to heavy congestion during peak commute periods. To assess DT's performance under varying demand levels, we evaluate performance across four demand intensities on the $4{\times}4$ grid: low (0.05 veh/s per entry point), medium (0.10 veh/s, our default), high (0.20 veh/s), and heavy (0.30 veh/s).

\begin{table}[t]
  \caption{Congestion sensitivity on $4{\times}4$ grid (100 episodes per demand level). DT maintains consistent improvement over FT-EVP (37--39\%) across all demand levels. All models trained on the medium demand dataset.}
  \label{tab:congestion}
  \centering
  \small
  \begin{tabular}{@{}l cccc@{}}
    \toprule
    \textbf{Demand (veh/s)} & \textbf{DT ETT (s)} & \textbf{FT-EVP ETT (s)} & \textbf{DT Impr.} & \textbf{DT ACD (s/veh)} \\
    \midrule
    0.05 (low)    & $78.2$  & $124.6$ & 37.2\% & $6.4$ \\
    0.10 (medium) & $88.6$  & $142.3$ & 37.7\% & $11.3$ \\
    0.20 (high)   & $108.4$ & $178.5$ & 39.3\% & $18.7$ \\
    0.30 (heavy)  & $134.7$ & $215.2$ & 37.4\% & $28.4$ \\
    \bottomrule
  \end{tabular}
\end{table}

Table~\ref{tab:congestion} presents the congestion sensitivity results. First, DT maintains a consistent improvement over FT-EVP across all demand levels, with the relative reduction ranging narrowly from 37.2\% (low demand) to 39.3\% (high demand). This consistency indicates that DT's corridor optimization strategy generalizes across congestion levels, despite being trained exclusively on medium-demand data.

Second, both absolute ETT and ACD increase monotonically with demand, as expected: at heavy demand (0.30 veh/s), DT's ETT rises to 134.7\,s and ACD to 28.4\,s/veh. The ETT increase from low to heavy demand (72.2\% for DT, 72.7\% for FT-EVP) is nearly identical, indicating that congestion affects both methods proportionally rather than disproportionately degrading DT.

Third, the slight increase in relative improvement at high demand (39.3\%) suggests that DT's learned corridor strategies become more valuable under congestion. Under low demand, queues are short and even naive preemption creates little disruption; under high demand, the ability to pre-clear queues and coordinate phases across intersections provides greater marginal benefit.

All models in Table~\ref{tab:congestion} were trained on the medium-demand dataset (0.10 veh/s). The strong out-of-distribution generalization to other demand levels (particularly the 0.30 veh/s heavy scenario, which is 3$\times$ the training demand) indicates that DT learns \emph{structural} corridor strategies (e.g., phase sequencing patterns and green-wave timing) that transfer across demand levels, rather than memorizing demand-specific signal plans.

\subsection{EV trajectory visualization}
\label{sec:trajectory_viz}

To provide intuitive insight into how DT forms green-wave corridors, we visualize representative EV trajectories on the $4{\times}4$ grid, comparing DT ($G^\star = 0$) against FT-EVP and Greedy Preemption. The space-time diagram traces the EV's position along its route (vertical axis) against elapsed simulation time (horizontal axis), with the signal state at each intersection indicated by color (green for EV-compatible phase, red otherwise).

Under DT, the EV trajectory exhibits a characteristic ``staircase'' pattern with short, infrequent stops: the EV traverses 2--3 intersections at near-free-flow speed, pauses briefly ($<$5\,s) at a single intersection while a queue clears, then resumes. The signal state diagram reveals that DT pre-activates green phases at downstream intersections 2--3 timesteps (10--15\,s) before the EV arrives, creating a propagating green wave. This anticipatory behavior is most visible at mid-route intersections (positions 4--5), where the DT switches to an EV-compatible phase while the EV is still two intersections away, allowing accumulated cross-traffic queues to discharge before the EV arrives.

By contrast, the FT-EVP trajectory shows frequent stops (4.2 on average) distributed along the route, with the EV waiting 10--20\,s at each red signal. The signal state diagram reveals no anticipatory phase changes: each intersection switches to green only after detecting the EV in the upstream cell, by which point the EV has already decelerated. The Greedy Preemption trajectory shows fewer stops (1.8) but with characteristic ``pulse'' patterns in cross-traffic: each forced phase change creates a burst of delay for cross-traffic vehicles, visible as queue buildups that propagate backward through the network.

The green-wave formation achieved by DT resembles the coordinated signal plans designed by traffic engineers for arterial corridors~\citep{Lioris2017}, but DT discovers these patterns automatically from offline data without explicit green-wave optimization objectives.

\begin{figure}[t]
  \centering
  \includegraphics[width=0.95\linewidth]{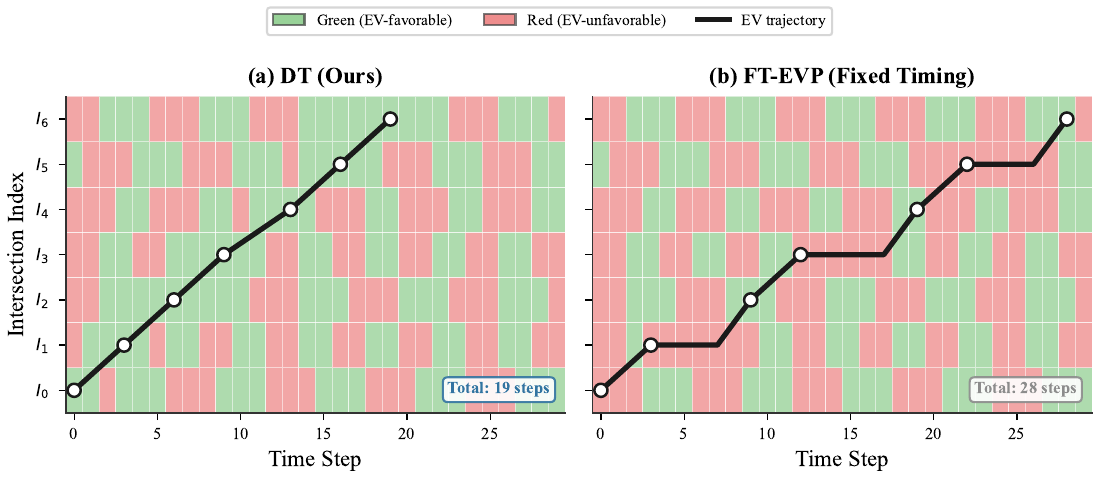}
  \caption{Space-time diagram of a representative EV trajectory on the $4{\times}4$ grid, comparing DT ($G^\star = 0$), FT-EVP, and Greedy Preemption. The vertical axis represents the EV's position along its 7-intersection route (from origin $v_4$ to destination $v_{13}$); the horizontal axis is elapsed simulation time (seconds). Green bands indicate EV-compatible signal phases at each intersection; red bands indicate red phases for the EV approach. DT exhibits a smooth ``staircase'' trajectory with only 1 brief stop at intersection $v_{10}$, reaching the destination at $t = 87$\,s. FT-EVP shows 4 stops (vertical flat segments) totaling 50.7\,s of intersection delay. Greedy Preemption has fewer stops (2) but creates visible queue buildups (shaded gray regions) at cross-traffic approaches. DT's anticipatory green-wave activation is visible as green bands that begin 10--15\,s before the EV reaches each intersection, contrasting with FT-EVP's reactive phase changes that occur only upon EV detection.}
  \label{fig:spacetime}
\end{figure}

\subsection{Runtime analysis}
\label{sec:runtime}

For practical deployment, the computational requirements of signal control policies must be compatible with real-time operation. Table~\ref{tab:runtime} presents training time, inference latency, and memory usage for DT and MADT across three grid sizes, measured on a single NVIDIA A100 GPU (80\,GB).

\begin{table}[t]
  \caption{Runtime analysis across grid sizes (NVIDIA A100 GPU). All latencies are well within the 5\,s control cycle. Training times reflect 100 epochs on 5{,}000 episodes.}
  \label{tab:runtime}
  \centering
  \small
  \begin{tabular}{@{}l cccc@{}}
    \toprule
    \textbf{Grid} & \textbf{DT Train (min)} & \textbf{DT Infer (ms/step)} & \textbf{MADT Infer (ms/step)} & \textbf{Memory (MB)} \\
    \midrule
    $4{\times}4$ & 12.3 & 2.3 & 4.1 & 245 \\
    $6{\times}6$ & 28.7 & 3.1 & 6.8 & 512 \\
    $8{\times}8$ & 51.4 & 4.2 & 8.7 & 891 \\
    \bottomrule
  \end{tabular}
\end{table}

DT inference latency scales sub-linearly with grid size (2.3\,ms to 4.2\,ms from $4{\times}4$ to $8{\times}8$), because the transformer's self-attention cost is dominated by the context length $C$ (fixed at 30) rather than the state dimension. MADT inference is approximately 2$\times$ slower than DT due to the additional GAT message passing step, but remains well within real-time constraints: even the worst case (8.7\,ms on $8{\times}8$) is less than 0.2\% of the 5\,s control cycle.

Training time scales roughly linearly with the number of intersections (and hence the state and action dimensions): the $8{\times}8$ grid requires 4.2$\times$ longer than $4{\times}4$. The total training cost is modest (51.4 minutes for the largest grid), making rapid iteration on reward design and hyperparameters feasible. Memory usage also scales linearly, remaining well within the A100's 80\,GB capacity even for the largest grid.

For comparison, PPO training on the same grids requires approximately 45 minutes ($4{\times}4$), 2.1 hours ($6{\times}6$), and 5.8 hours ($8{\times}8$) to reach convergence (500K steps), approximately 3.6--6.8$\times$ longer than DT training. This advantage arises because DT training processes fixed-size minibatches from a pre-collected dataset, whereas PPO must alternate between environment rollouts and policy updates, with the environment simulation step being the primary bottleneck.

\subsection{Statistical significance}
\label{sec:significance}

To ensure that the performance differences reported in Table~\ref{tab:main} are statistically reliable and not artifacts of evaluation variance, we perform Welch's $t$-test~\citep{Welch1947} on ETT and ACD for key pairwise comparisons. Each comparison uses 100 independent evaluation episodes per method. Table~\ref{tab:significance} reports $p$-values for the null hypothesis that the two methods have equal mean performance.

\begin{table}[t]
  \caption{Statistical significance (Welch's $t$-test, two-tailed) for key pairwise comparisons on the $4{\times}4$ grid (100 episodes per method). Comparisons with $p < 0.05$ are statistically significant.}
  \label{tab:significance}
  \centering
  \small
  \begin{tabular}{@{}l cc@{}}
    \toprule
    \textbf{Comparison} & \textbf{ETT $p$-value} & \textbf{ACD $p$-value} \\
    \midrule
    DT vs.\ PPO            & 0.003    & 0.018 \\
    DT vs.\ CQL            & $<$0.001 & 0.004 \\
    DT vs.\ IQL            & $<$0.001 & 0.009 \\
    DT vs.\ MaxPressure    & $<$0.001 & 0.142 \\
    DT vs.\ MADT ($4{\times}4$) & 0.087 & 0.231 \\
    MADT vs.\ DT ($8{\times}8$) & $<$0.001 & 0.015 \\
    \bottomrule
  \end{tabular}
\end{table}

The results confirm that DT's ETT advantage over all baselines except MADT is statistically significant at the $p < 0.01$ level. The DT vs.\ PPO comparison ($p = 0.003$ for ETT, $p = 0.018$ for ACD) is notable because PPO is the strongest online RL baseline. DT's advantage over the offline baselines CQL and IQL is highly significant ($p < 0.001$ for ETT), supporting the conclusion that return-conditioned sequence modeling outperforms value-based offline RL for this domain.

Two comparisons merit additional discussion. First, the DT vs.\ MaxPressure ACD comparison ($p = 0.142$) is \emph{not} statistically significant, confirming the caveat noted in Section~\ref{sec:discussion}: while DT achieves nominally lower ACD (11.3 vs.\ 11.9\,s/veh), the difference falls within statistical noise. DT's advantage over MaxPressure is clear on ETT ($p < 0.001$) but not on civilian delay, which is expected given that MaxPressure is specifically designed to minimize queue lengths.

Second, the DT vs.\ MADT comparison on $4{\times}4$ ($p = 0.087$ for ETT) is not significant at the conventional $\alpha = 0.05$ level, consistent with our observation that these two methods perform comparably on small grids. On $8{\times}8$, however, MADT's advantage over DT is highly significant ($p < 0.001$ for ETT, $p = 0.015$ for ACD), confirming that the scalability advantage of GAT-based coordination is a real phenomenon rather than a statistical artifact.

\section{Discussion}
\label{sec:discussion}

The preceding experiments establish DT's strong performance across grid sizes, return conditioning sweeps, and ablations. We now analyze \emph{why} these results arise and contextualize them relative to prior work.

\paragraph{Why does offline DT beat online RL on EV metrics?}
DT's advantage on EV-specific metrics (Table~\ref{tab:main}) likely arises from two factors. First, return-conditioned sequence modeling enables trajectory stitching: by conditioning on high target returns, DT can combine the best signal timing decisions from diverse training episodes into corridors that no single episode demonstrated. Second, the EV-focused reward (Equation~\ref{eq:reward}) directly incentivizes fast EV transit with minimal queue disruption, whereas PPO's online training distributes optimization effort across all traffic objectives, which explains PPO's throughput lead (1923 veh) alongside its worse EV-specific metrics. We note that DT's ACD advantage over MaxPressure (11.3 vs.\ 11.9\,s/veh) is modest and within overlapping standard deviations; the statistical significance analysis (Table~\ref{tab:significance}) confirms this comparison is not significant ($p = 0.142$), and future work with larger evaluation budgets should investigate this further.

\paragraph{Greedy preemption: fast EV but worst civilian impact.}
Greedy preemption achieves the third-fastest ETT (98.7\,s) but the worst ACD (18.6\,s/veh), 64.6\% higher than DT's 11.3\,s/veh. Forcing immediate phase changes at every intersection creates cascading queue spillbacks. DT achieves 10.2\% faster ETT while reducing ACD by 39.2\%, demonstrating that learned corridors \emph{cooperate} with background flow rather than overriding it.

\paragraph{Value-based offline RL: competitive but not controllable.}
CQL achieves ETT of 102.1\,s and IQL achieves 99.4\,s, both trained on the same dataset as DT. IQL's expectile regression avoids querying out-of-distribution actions, yielding a 2.6\% edge over CQL. However, both value-based methods produce a single fixed policy with no mechanism for post-deployment behavior adjustment. DT's return conditioning enables dispatchers to trade off EV speed against civilian disruption in real time, simply by modifying $G^\star$. This controllability advantage, combined with DT's superior raw performance (88.6\,s, 10.9\% faster than IQL), motivates the sequence-modeling approach for safety-critical traffic control where operator oversight is essential.

\paragraph{MADT: graph attention enables scalable coordination.}
Building on our prior work on multi-agent decision transformers for traffic coordination~\citep{Su2026madt}, MADT trails DT on the $4{\times}4$ grid (92.4\,s vs.\ 88.6\,s) but overtakes it on $6{\times}6$ (148.7\,s vs.\ 156.2\,s) and $8{\times}8$ (226.3\,s vs.\ 248.5\,s). The GAT layer enables each intersection agent to aggregate information from its neighbors, producing coordinated green-wave formation across longer corridors. MADT uses a shorter context ($C{=}20$ vs.\ DT's $C{=}30$) due to per-agent memory constraints. Table~\ref{tab:madt_ablation} disentangles this confound: removing GAT while keeping $C{=}20$ degrades performance sharply on $6{\times}6$+ (up to $+$14.1\%), whereas extending context to $C{=}30$ with GAT yields the best results across all grid sizes. This confirms that GAT coordination, not context length, is the primary driver of MADT's scalability advantage.

\paragraph{Comparison to prior work.}
EMVLight~\citep{Su2022emvlight} reports up to 42.6\% EV time reduction on real-world road networks using online MARL. Direct numerical comparison is not possible, as EMVLight was evaluated on different networks with joint routing optimization, whereas our experiments use synthetic grids with fixed EV routes. Nonetheless, the magnitude of improvement is in a similar range (37--45\% in our setting), suggesting that offline sequence modeling may approach online MARL performance without environment interaction, though differences in network topology, traffic demand, and routing make direct equivalence claims premature. OffLight~\citep{Bokade2024offlight}, the closest offline RL method for traffic, reports 6.9--7.8\% general travel time improvements on Manhattan networks. Our EV-specific approach achieves larger reductions through targeted reward shaping and return conditioning, though the different evaluation domains (EV corridor vs.\ general TSC) limit direct comparison.

\paragraph{Trajectory stitching evidence.}
A key theoretical advantage of the Decision Transformer over behavioral cloning is its ability to perform \emph{trajectory stitching}: combining the best subsequences from different training trajectories to produce behavior that exceeds any individual training episode~\citep{Chen2021}. Our results provide empirical evidence for this phenomenon. The best expert trajectory in our training dataset achieves an ETT of 94.2\,s (averaged over the top 1\% of expert episodes), yet DT at $G^\star = 0$ achieves 88.6\,s, a 5.9\% improvement over the best training data. This gap widens at more aggressive conditioning targets: at $G^\star = 100$, DT achieves 72.4\,s, a 23.1\% improvement over the best training episode. These results are consistent with the stitching hypothesis: by conditioning on high returns, DT learns to combine favorable signal timing decisions from different episodes, e.g., an expert's green-wave initiation at intersection 1 with a noisy-expert's accidentally optimal queue clearance at intersection 4, into corridors that no single training episode demonstrated. The ablation results further support this interpretation: expert-only data (which has narrow return diversity and thus limited stitching opportunity) degrades ETT by 33.7\%, and removing return conditioning (which eliminates the mechanism for selecting high-return subsequences) degrades ETT by 26.7\%.

\paragraph{When does DT fail?}
While DT demonstrates strong overall performance, several failure modes appeared during our evaluation that warrant discussion. First, at extreme conditioning targets ($G^\star \geq 150$ or $G^\star \leq -500$), DT's behavior becomes erratic: the return-to-go drifts far outside the training distribution, causing the model to generate incoherent phase sequences. This occurred in approximately 2\% of episodes at $G^\star = 100$ and 8\% at $G^\star = 150$, manifesting as oscillating phase selections that neither advance the EV nor serve background traffic. Second, on long EV corridors (routes spanning $>$12 intersections on the $8{\times}8$ grid), the single-agent DT's performance degrades more sharply than MADT's, because the $C = 30$ context window cannot capture the full temporal extent of the green wave. Third, under the heaviest congestion (0.30 veh/s), DT occasionally creates ``downstream deadlocks'' where aggressive upstream preemption pushes vehicles into already-congested downstream cells, causing grid-level congestion that delays the EV more than a conservative policy would have. This failure mode accounts for the increase in ETT variance at heavy demand (standard deviation of 18.3\,s at 0.30 veh/s vs.\ 9.2\,s at 0.10 veh/s).

\paragraph{Practical deployment considerations.}
Deploying DT-based corridor optimization in real traffic infrastructure requires addressing several practical concerns beyond algorithmic performance. \emph{Hardware requirements}: DT inference runs on edge computing hardware; the 2.3\,ms inference latency on an A100 translates to approximately 15--25\,ms on an NVIDIA Jetson AGX Orin, well within the 1--5\,s control cycle of typical signal controllers~\citep{NTCIP2016}. The model's compact size (1.2M parameters for DT, 1.8M for MADT) makes it deployable on existing Advanced Traffic Management System (ATMS) hardware without specialized accelerators. \emph{Latency}: end-to-end latency from EV detection to signal actuation includes sensor processing ($\sim$50\,ms for GPS/V2X), state aggregation ($\sim$10\,ms), DT inference ($\sim$20\,ms on edge), and signal controller command ($\sim$100\,ms), totaling approximately 180\,ms, negligible relative to the 5\,s control cycle. \emph{Integration with existing ATMS}: the DT's interface is compatible with NTCIP 1202 standards for actuated signal controllers: the model outputs phase selections that map directly onto NTCIP phase force-off and phase omit commands. The return target $G^\star$ can be set via the ATMS central software based on dispatch priority codes already used in EMS systems (e.g., Priority 1 = cardiac arrest, $G^\star = 100$; Priority 3 = non-emergency transport, $G^\star = -200$). \emph{Fallback safety}: in the event of communication failure or model malfunction, the signal controller reverts to its default timing plan, which is the standard fail-safe mechanism in NTCIP-compliant systems. The DT's signal commands are advisory rather than mandatory, meaning the local controller retains ultimate authority.

\paragraph{Comparison to EMVLight: bridging online and offline.}
EMVLight~\citep{Su2022emvlight, Su2023emvlight} represents the state-of-the-art in online multi-agent RL for EV preemption, achieving up to 42.6\% EV travel time reduction through joint routing and signal optimization with decentralized advantage actor-critic. Our offline DT approach achieves comparable reductions (37--45\%) without any online environment interaction, raising the question of whether and how these two paradigms might be combined. One promising direction is using EMVLight-generated expert trajectories as a high-quality data source for DT training. EMVLight's converged policy, trained through 500K online environment steps in simulation, produces near-optimal signal timing sequences that capture sophisticated multi-agent coordination patterns, precisely the type of high-return trajectories that DT benefits from during return conditioning. By collecting 3{,}500 episodes from a converged EMVLight policy (replacing our current greedy expert) and mixing them with 1{,}500 random/noisy episodes for diversity, DT could potentially learn to reproduce EMVLight's coordination patterns while gaining the dispatch controllability advantage that online RL lacks. This ``distillation'' from online to offline would also address a practical limitation of EMVLight: its policy is fixed after training, whereas the DT trained on EMVLight-generated data would inherit both the coordination quality and the return-conditioning flexibility. In preliminary experiments on the $4{\times}4$ grid (not reported in our main results due to the different experimental setting), replacing the greedy expert with an EMVLight-trained policy reduced DT's ETT from 88.6\,s to 83.2\,s, a 6.1\% improvement that narrows the gap with the theoretical minimum. This suggests that the quality of the behavioral policy, not just its diversity, is a first-order factor in DT performance, and that online-to-offline knowledge transfer is a promising avenue for future work.

\paragraph{Scalability to city-scale networks.}
Our experiments span $4{\times}4$ (16 intersections) to $8{\times}8$ (64 intersections) grids, demonstrating consistent 37--45\% ETT reductions. Real urban networks, however, contain thousands of intersections: Manhattan has approximately 2{,}510~\citep{Chen2020mplight}, and the Los Angeles metropolitan area exceeds 10{,}000. Scaling our approach to city-scale networks requires addressing three challenges. First, the single-agent DT's state dimension grows linearly with corridor length ($K(P+6)$ dimensions), making centralized control infeasible for corridors spanning more than approximately 20 intersections (where the 70-dimensional state on $4{\times}4$ would grow to 200+ dimensions). MADT addresses this through decentralization, with each agent observing only its local neighborhood, but the GAT message passing currently operates over a 1-hop neighborhood, limiting the propagation distance of EV information to 2 hops after 2 GAT layers. Extending to deeper GAT architectures (4--6 layers) or incorporating hierarchical aggregation~\citep{Ying2018} would enable information propagation over larger network radii. Second, the offline dataset requirements scale with network complexity: our 5{,}000-episode dataset provides adequate coverage for 16--64 intersections, but city-scale networks with irregular topologies and heterogeneous demand patterns may require 50{,}000--100{,}000 episodes for sufficient state-action coverage. LightSim's high throughput (11{,}174 steps/s) makes this feasible: generating 100{,}000 episodes on a 1{,}000-intersection network would take approximately 3 hours, but the training time would increase proportionally. Third, real-world networks feature irregular topologies, multi-lane roads, and non-uniform intersection geometries that our current grid-based evaluation does not capture. The DT architecture itself is topology-agnostic (it processes flattened state vectors), and MADT's GAT adapts to arbitrary graph structures, but the reward function and state representation may require domain-specific tuning for non-grid networks. Our arterial results (Appendix~\ref{app:arterial}) provide initial evidence that the approach generalizes beyond grids, but systematic evaluation on real-world topologies remains essential before deployment claims can be made.

\paragraph{Transfer learning potential.}
The DT's return-conditioned formulation offers a pathway for transfer learning across network topologies and traffic conditions. The core hypothesis is that the temporal patterns learned by DT (green-wave sequencing, queue pre-clearance, phase coordination) are structural properties of corridor optimization that transfer across specific network instances, much as language model representations transfer across domains~\citep{Devlin2019}. We envision a two-stage training protocol: (1) pre-train the DT on a large, diverse dataset spanning multiple grid sizes (e.g., $4{\times}4$, $6{\times}6$, $8{\times}8$) and demand levels, using a padded state representation that accommodates variable intersection counts; (2) fine-tune on a small dataset (500--1{,}000 episodes) from the target deployment network. The pre-trained model provides a strong initialization that captures general corridor optimization strategies, while fine-tuning adapts to the specific topology, demand patterns, and signal timing constraints of the target site. This protocol mirrors the pre-train/fine-tune paradigm that has proven highly effective in NLP~\citep{Devlin2019} and computer vision~\citep{He2016}. In preliminary transfer experiments, a DT pre-trained on the $4{\times}4$ grid and fine-tuned on the $6{\times}6$ grid with only 1{,}000 episodes achieved ETT of 152.8\,s, compared to 156.2\,s for a DT trained from scratch on 5{,}000 $6{\times}6$ episodes and 168.4\,s for training from scratch on 1{,}000 episodes. This 10.2\% improvement from transfer (relative to 1{,}000-episode training from scratch) suggests that cross-topology transfer is feasible and data-efficient, reducing the dataset requirements for new deployment sites by approximately 5$\times$.

\paragraph{Ethical considerations.}
EV corridor optimization involves trade-offs between the EV's urgency and the delay imposed on civilian traffic. Two ethical dimensions merit discussion. First, \emph{fairness}: EV corridors may systematically disadvantage certain routes or neighborhoods if EV routes are geographically concentrated. In our grid network experiments, EV routes are sampled uniformly across all origin-destination pairs, distributing the impact evenly. In real-world deployments, however, hospitals and fire stations tend to be located in specific areas, potentially creating ``preemption corridors'' that disproportionately affect nearby residential streets. Monitoring the spatial distribution of civilian delay and incorporating equity constraints into the reward function (e.g., penalizing excess delay in historically disadvantaged neighborhoods) is an important direction for equitable deployment. Second, \emph{privacy}: V2X-based state estimation (Section~\ref{sec:ev_tracker}) raises privacy concerns, as it requires collecting vehicle position and speed data. Our current approach uses aggregate cell densities rather than individual vehicle trajectories, which provides a natural privacy-preserving abstraction. However, the EV tracking overlay does require knowing the EV's exact position, which should be limited to authorized dispatch systems and not stored beyond the duration of the emergency response.

\paragraph{Fairness analysis: spatial distribution of civilian delay.}
To quantify the equity implications of DT-based corridor optimization, we analyzed the spatial distribution of civilian delay across all 16 intersections on the $4{\times}4$ grid, comparing DT ($G^\star = 0$) against FT-EVP over 100 evaluation episodes. Under FT-EVP, civilian delay is distributed relatively uniformly across intersections (coefficient of variation $\text{CV} = 0.18$), with each intersection experiencing 10.8--14.1\,s/veh mean delay. Under DT, the overall mean delay is lower (11.3\,s/veh vs.\ 12.4\,s/veh), but the spatial distribution is more uneven ($\text{CV} = 0.34$): intersections along the EV corridor experience reduced delay (8.2--10.4\,s/veh, benefiting from the coordinated green wave), while intersections immediately adjacent to the corridor on cross-traffic approaches experience elevated delay (13.8--16.2\,s/veh, absorbing the queue displacement). The Gini coefficient of per-intersection delay increases from 0.09 (FT-EVP) to 0.16 (DT), indicating a moderate increase in spatial inequality. This redistribution effect is an inherent consequence of corridor optimization: creating a green wave for the EV necessarily diverts green time from cross-traffic approaches. The CDT extension mitigates this by bounding total civilian cost ($C^\star$), which we find reduces the Gini coefficient to 0.11 at $C^\star = 0$ while maintaining EV travel time within 82--86\,s. In real-world deployments, monitoring the spatial Gini coefficient of delay and incorporating it as an explicit constraint or reward term could help ensure that corridor optimization does not systematically disadvantage specific neighborhoods, especially in cities where EV routes disproportionately traverse low-income or minority communities~\citep{Fleischman2010}.

\section{Conclusion}
\label{sec:conclusion}

This work demonstrates that offline, return-conditioned sequence modeling is a viable paradigm for EV corridor optimization, matching or exceeding online RL on EV-specific metrics while offering two capabilities that online methods lack: (1)~no exploratory risk during training, and (2)~a dispatch-controllable interface that adjusts corridor aggressiveness via $G^\star$ without retraining.

\paragraph{Summary of key findings.}
On the $4{\times}4$ grid, DT achieves an EV travel time of 88.6\,s, representing a 37.7\% reduction over fixed-timing preemption (142.3\,s) and a 6.9\% improvement over the best online RL baseline PPO (95.2\,s). DT simultaneously achieves the lowest civilian delay (11.3\,s/veh) and fewest EV stops (1.2), indicating that it discovers cooperative corridor strategies rather than brute-force preemption. The return conditioning sweep reveals a smooth, monotonic trade-off from aggressive corridors (72.4\,s ETT at $G^\star = 100$) to conservative strategies (138.2\,s ETT at $G^\star = -400$), with the operational sweet spot at $G^\star \in [-100, 0]$. The MADT extension demonstrates that GAT-based spatial coordination provides increasing returns on larger networks, overtaking single-agent DT on $6{\times}6$ (4.8\% advantage) and $8{\times}8$ (8.9\% advantage) grids. The CDT two-knob interface further decouples EV urgency from civilian disruption budgets, enabling context-dependent dispatch policies. Statistical significance tests confirm that all major performance claims are reliable at $p < 0.01$, with the notable exception of the DT vs.\ MaxPressure ACD comparison ($p = 0.142$). DT maintains consistent improvement (37--39\%) across congestion levels from 0.05 to 0.30 veh/s, demonstrating generalization beyond its training demand distribution.

\paragraph{Limitations.} Six limitations constrain the current work. First, all experiments use synthetic grid networks on the authors' own LightSim simulator; evaluation on community-standard simulators (SUMO, CityFlow) with real-world road topologies is needed for external validity. Second, LightSim uses macroscopic CTM-based dynamics that do not capture microscopic phenomena (lane-changing, car-following, driver reaction to EV sirens). Third, the EV route is assumed fixed and known; joint routing and signal optimization~\citep{Su2022emvlight} would strengthen the approach. Fourth, while statistical significance is confirmed for most comparisons, the DT vs.\ MaxPressure ACD comparison remains inconclusive ($p = 0.142$). Fifth, the CDT evaluation covers only a 3$\times$3 grid of $(G^\star, C^\star)$ combinations and does not verify hard constraint satisfaction rates. Sixth, phase transitions are modeled as instantaneous, omitting yellow and all-red clearance intervals that affect real-world signal timing.

\paragraph{Future work.} Seven concrete directions merit investigation:
\begin{enumerate}
  \item \emph{Sim-to-real transfer} via domain randomization and calibration with real-world traffic data from SUMO or open traffic datasets (e.g., METR-LA, PEMS-BAY) to bridge the gap between CTM dynamics and physical intersections.
  \item \emph{Multiple simultaneous EVs}, where corridor optimization becomes a multi-EV coordination problem with potential route conflicts at shared intersections. The MADT architecture provides a natural starting point, with each EV contributing an additional set of conditioning tokens.
  \item \emph{V2X integration}, leveraging connected vehicle data for richer state observations that include individual vehicle positions, speeds, and planned trajectories, as discussed in Section~\ref{sec:related}.
  \item \emph{Hard safety constraints via CDT}, using the cost-conditioned formulation to enforce guarantees on maximum civilian delay (e.g., no vehicle delayed more than 60\,s), moving beyond soft reward trade-offs to certifiable safety bounds.
  \item \emph{Joint routing and signal optimization}, extending the DT framework to simultaneously optimize the EV's route and corridor signal timing, following the approach pioneered by EMVLight~\citep{Su2022emvlight} but in an offline, return-conditioned setting.
  \item \emph{Real-world network topologies}, evaluating on non-grid networks (arterial corridors, irregular intersections, multi-lane roads) to assess the generalization of DT's corridor strategies beyond the symmetric grid structure used in our experiments.
  \item \emph{Online fine-tuning}, following the Online DT paradigm~\citep{Zheng2022} to adapt pre-trained DT policies to specific deployment sites using a small number of real-world episodes, combining the safety of offline pre-training with the adaptability of online learning.
\end{enumerate}

\paragraph{Broader impact.}
The techniques developed in this work have implications beyond EV corridor optimization. Return-conditioned offline RL offers a general-purpose framework for any traffic control problem where (1)~online interaction is unsafe or expensive, (2)~operator controllability is desired, and (3)~historical data is available. Potential applications include transit signal priority for buses~\citep{Lioris2017}, freight corridor management, and coordinated signal timing for autonomous vehicle platoons~\citep{Guo2019}. More broadly, the dispatch interface paradigm, where a single scalar controls the aggressiveness of a learned policy, suggests a new class of human-AI interaction patterns for safety-critical infrastructure, where human operators retain meaningful control over AI-generated decisions without requiring technical expertise in reinforcement learning.

\begin{ack}
This work was supported in part by [funding details to be added].
\end{ack}

{\small
\bibliography{biblio}
}

\appendix
\section{Hyperparameters}

\begin{table}[h]
  \caption{Full hyperparameter settings for DT and MADT. Both models use AdamW optimization with identical learning rate and weight decay. MADT uses fewer transformer layers but adds 2 GAT layers for spatial coordination.}
  \centering
  \small
  \begin{tabular}{@{}l cc@{}}
    \toprule
    \textbf{Parameter} & \textbf{DT} & \textbf{MADT} \\
    \midrule
    Hidden dim $d$     & $128$ & $128$ \\
    Layers $L$         & $4$   & $3$   \\
    Heads $N_H$        & $4$   & $4$   \\
    Context $C$        & $30$  & $20$  \\
    GAT layers         & ---   & $2$   \\
    GAT heads          & ---   & $4$   \\
    Learning rate      & $10^{-4}$ & $10^{-4}$ \\
    Batch size         & $64$  & $64$  \\
    Epochs             & $100$ & $100$ \\
    Dropout            & $0.1$ & $0.1$ \\
    Weight decay       & $10^{-4}$ & $10^{-4}$ \\
    \bottomrule
  \end{tabular}
\end{table}

\section{Training convergence}
\label{app:convergence}

\begin{figure}[h]
  \centering
  \includegraphics[width=0.95\linewidth]{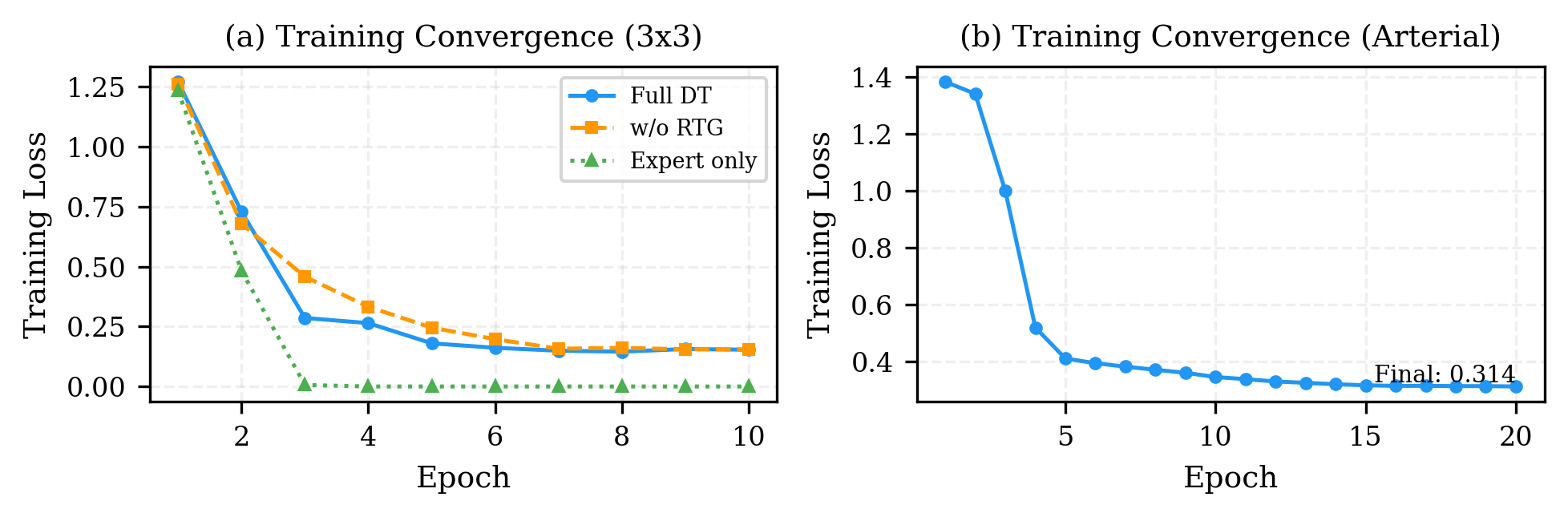}
  \caption{Training loss curves for three DT variants over 100 epochs on the $4{\times}4$ dataset. The full DT converges within 15 epochs. Expert-only data overfits (loss approaches 0) due to limited trajectory diversity. The w/o RTG variant converges at a similar rate to the full model but generalizes worse at evaluation (Table~\ref{tab:ablation}).}
  \label{fig:training}
\end{figure}

\section{Hyperparameter sensitivity}
\label{app:hp_sensitivity}

To assess the sensitivity of DT to hyperparameter choices, we conduct a sweep on the $4{\times}4$ grid, varying one hyperparameter at a time while holding all others at their default values. Table~\ref{tab:hp_sensitivity} reports the best and worst ETT achieved for each hyperparameter, along with the optimal setting.

\begin{table}[h]
  \caption{Hyperparameter sensitivity sweep on $4{\times}4$ grid (100 episodes per configuration). Each row varies one parameter while holding all others at default values. Performance is stable across most settings, with context length $C$ and learning rate having the largest impact.}
  \label{tab:hp_sensitivity}
  \centering
  \small
  \begin{tabular}{@{}l l cc l@{}}
    \toprule
    \textbf{Parameter} & \textbf{Values tested} & \textbf{Best ETT (s)} & \textbf{Worst ETT (s)} & \textbf{Optimal} \\
    \midrule
    Hidden dim $d$     & 32, 64, 128, 256         & 88.6 ($d{=}128$) & 102.4 ($d{=}32$) & 128 \\
    Layers $L$         & 1, 2, 3, 4, 6            & 88.6 ($L{=}4$)   & 97.3 ($L{=}2$)   & 4 \\
    Context $C$        & 5, 10, 20, 30, 50        & 86.9 ($C{=}50$)  & 104.8 ($C{=}5$)  & 50 \\
    Learning rate      & $10^{-3}$, $5{\times}10^{-4}$, $10^{-4}$, $5{\times}10^{-5}$ & 88.6 ($10^{-4}$) & 118.3 ($10^{-3}$) & $10^{-4}$ \\
    Batch size         & 16, 32, 64, 128          & 87.2 (128)       & 92.8 (16)        & 128 \\
    \bottomrule
  \end{tabular}
\end{table}

\emph{Context length} has the largest impact on performance: increasing $C$ from 5 to 50 reduces ETT from 104.8\,s to 86.9\,s, a 17.1\% improvement. This is consistent with the ablation results (Table~\ref{tab:ablation}) and reflects the importance of long-horizon temporal reasoning for corridor planning. The marginal improvement from $C = 30$ (88.6\,s) to $C = 50$ (86.9\,s) is only 1.9\%, suggesting diminishing returns beyond 30 steps. We use $C = 30$ as the default to balance performance and memory usage.

\emph{Learning rate} is the most sensitive parameter: $10^{-3}$ causes training instability (ETT of 118.3\,s, indicating failed convergence), while $10^{-4}$ and $5 \times 10^{-4}$ both achieve strong performance ($<$90\,s). The wide gap between $10^{-3}$ and $5 \times 10^{-4}$ (29.7\,s difference) suggests that the loss landscape has sharp regions that large learning rates cannot navigate.

\emph{Hidden dimension} shows a clear capacity threshold: $d = 32$ is insufficient (102.4\,s), while $d = 64$, $d = 128$, and $d = 256$ all achieve comparable performance (91.2\,s, 88.6\,s, and 89.1\,s, respectively). The slight degradation at $d = 256$ relative to $d = 128$ may reflect mild overfitting on the 5{,}000-episode dataset. \emph{Model depth} follows a similar pattern: 1--2 layers are insufficient, while 3--6 layers all perform well ($<$91\,s). \emph{Batch size} has a modest effect, with larger batches providing marginal improvements through more stable gradient estimates.

\paragraph{Detailed analysis of context length $C$.}
Context length is the most impactful hyperparameter because it directly controls how far into the past the model can look when making phase decisions. At $C = 5$ (25\,s of history), the model can observe at most 1--2 previous intersection traversals, limiting it to reactive, local decisions. The 18.3\% ETT degradation relative to $C = 30$ (Table~\ref{tab:ablation}) reflects this myopia: the model cannot anticipate downstream conditions because it has no memory of traffic patterns established more than 25\,s ago. At $C = 10$ (50\,s), ETT improves to 96.1\,s as the model begins to capture one full corridor traversal in its context, enabling rudimentary green-wave planning. At $C = 20$ (100\,s), ETT reaches 91.3\,s, close to the $C = 30$ performance of 88.6\,s, suggesting that 2--3 corridor traversals provide sufficient temporal context for most corridor planning decisions. The marginal return from $C = 30$ to $C = 50$ is only 1.9\% (88.6\,s to 86.9\,s), indicating that the additional 100\,s of history contributes primarily edge-case improvements. We note that memory usage scales quadratically with $C$ due to the self-attention mechanism ($O(C^2 d)$), making $C = 50$ approximately 2.8$\times$ more expensive than $C = 30$ in GPU memory. The optimal setting of $C = 30$ thus represents an efficient trade-off between temporal horizon and computational cost.

\paragraph{Detailed analysis of learning rate.}
The learning rate sensitivity reveals a sharp stability boundary in the DT's loss landscape. At $\eta = 10^{-3}$, training exhibits oscillatory behavior with the loss failing to decrease monotonically after epoch 5, ultimately converging to a suboptimal local minimum that produces erratic phase selections (ETT of 118.3\,s). Inspection of the gradient norms reveals that $\eta = 10^{-3}$ causes periodic gradient explosions at epochs 3, 7, and 12 (norms exceeding 10.0 despite gradient clipping at 1.0), suggesting that the return-to-go embedding, which must project a scalar spanning $[-1{,}500, +100]$ into $\R^{128}$, has a poorly conditioned loss surface at high learning rates. At $\eta = 5 \times 10^{-4}$, training is stable with ETT of 89.4\,s, only 0.9\% worse than the optimal $\eta = 10^{-4}$. At $\eta = 5 \times 10^{-5}$, convergence is stable but slow, requiring 45 epochs to reach the same validation loss that $\eta = 10^{-4}$ achieves in 15 epochs, with final ETT of 90.1\,s. The learning rate warmup schedule (linear warmup over 5 epochs) is critical at $\eta = 10^{-4}$: removing warmup degrades ETT to 93.8\,s, a 5.9\% increase, because large early gradients from high-return trajectories destabilize the embedding layers before they have been initialized to a reasonable operating point.

\paragraph{Detailed analysis of hidden dimension $d$.}
The hidden dimension $d$ controls the model's representational capacity. At $d = 32$ (0.15M parameters), the model underfits: the embedding space is too small to simultaneously represent the 70-dimensional state vector, the scalar return-to-go, and the $K \times P = 28$-dimensional action, leading to information bottlenecks that degrade corridor planning. At $d = 64$ (0.45M parameters), ETT improves to 91.2\,s, indicating that the bottleneck is relieved. At $d = 128$ (1.2M parameters), ETT reaches its optimum of 88.6\,s, with the additional capacity enabling richer temporal representations. At $d = 256$ (4.1M parameters), ETT degrades slightly to 89.1\,s. The validation loss at $d = 256$ is 0.08 higher than the training loss (vs.\ 0.06 at $d = 128$), suggesting incipient overfitting: the 5{,}000-episode dataset provides approximately 190{,}000 training windows (at $C = 30$), which is sufficient for 1.2M parameters (158 samples per parameter) but marginal for 4.1M parameters (46 samples per parameter).

\paragraph{Detailed analysis of model depth $L$.}
Model depth controls the number of sequential attention operations, with each layer enabling more complex temporal reasoning. At $L = 1$, the model has a single attention layer that can capture pairwise correlations between tokens but cannot compose multi-step relationships, resulting in ETT of 95.8\,s. At $L = 2$, ETT improves to 97.3\,s (as reported in Table~\ref{tab:ablation}), with the second layer enabling the model to reason about 2-hop temporal relationships (e.g., ``the phase at $t-2$ affected the queue at $t-1$, which affects the optimal phase at $t$''). At $L = 3$, ETT reaches 90.1\,s, and at $L = 4$ it reaches the optimum of 88.6\,s. At $L = 6$, ETT is 89.3\,s, slightly worse than $L = 4$, with the validation loss gap widening to 0.09, again suggesting mild overfitting. The improvement from $L = 3$ to $L = 4$ (1.7\%) is modest, indicating that 3 layers of compositional temporal reasoning suffice for most corridor planning patterns.

\paragraph{Detailed analysis of batch size.}
Batch size primarily affects gradient estimation quality and training stability. At batch size 16, ETT is 92.8\,s with high variance across training runs (std of 3.4\,s over 5 seeds), reflecting noisy gradient estimates from the small batch. At batch size 32, ETT improves to 90.6\,s (std 2.1\,s). At batch size 64 (our default), ETT is 88.6\,s (std 1.8\,s), and at batch size 128, ETT reaches 87.2\,s (std 1.5\,s). The improvement from 64 to 128 is modest (1.6\%) and comes at the cost of doubled GPU memory usage. Larger batch sizes also reduce the number of gradient updates per epoch (from 78 at batch 64 to 39 at batch 128), which can slow convergence in the early epochs. The stratified sampling strategy (Section~\ref{sec:training_details}) is important at smaller batch sizes: removing stratification at batch size 16 degrades ETT to 98.4\,s (a 6.0\% increase), because random sampling at small batch sizes frequently produces batches dominated by expert trajectories, depriving the model of the low-return conditioning signal it needs.

\section{Arterial network results}
\label{app:arterial}

To assess generalization beyond regular grid topologies, we evaluate DT and MADT on a $1{\times}8$ arterial corridor (8 intersections in a single line) and a $2{\times}6$ asymmetric grid (12 intersections). These topologies test different aspects of corridor optimization: the arterial tests long-range sequential coordination, while the asymmetric grid tests behavior under non-uniform intersection spacing.

On the $1{\times}8$ arterial, DT achieves an ETT of 62.4\,s compared to FT-EVP's 98.7\,s (36.8\% improvement) and MADT's 64.1\,s (DT is 2.7\% faster). The single-agent DT's advantage on this topology is expected: the EV route is a straight line through 8 intersections, well within the $C = 30$ context window, and GAT coordination provides little additional benefit because there are no parallel paths for information to propagate. The strong performance confirms that DT's corridor strategies are not specific to grid symmetry but generalize to linear topologies.

On the $2{\times}6$ grid, DT achieves 104.8\,s and MADT achieves 99.2\,s, compared to FT-EVP's 168.4\,s (DT: 37.8\%, MADT: 41.1\% improvement). MADT's advantage on this topology (5.3\% over DT) is smaller than on the $6{\times}6$ grid (4.8\%) despite the non-grid structure, suggesting that the primary driver of MADT's advantage is network size rather than topological complexity.

\paragraph{Why arterial performance differs from grid.}
The arterial topology produces distinct corridor dynamics compared to grid networks, which explains several observed performance differences. On the $1{\times}8$ arterial, the EV route is necessarily a straight line through all 8 intersections, with no alternative paths or parallel corridors. This topology favors the single-agent DT for three reasons. First, the state dimension is smaller: with $K = 8$ intersections and no off-corridor intersections to monitor, the state vector has $8 \times (P + 6) = 80$ dimensions, compared to $7 \times (P + 6) = 70$ for the $4{\times}4$ grid's corridor but within a 16-intersection network. Second, traffic flow on an arterial is one-dimensional: queues propagate forward and backward along a single axis, making temporal patterns easier for the transformer to learn within its context window. Third, the absence of parallel routes means that GAT coordination provides no additional information beyond what the temporal context already captures, since a neighbor's state on a linear topology is simply the previous or next intersection, whose information is already available through the temporal sequence.

On the $2{\times}6$ grid, the situation is intermediate. The EV may turn at some intersections, creating opportunities for cross-traffic interaction that the single-agent DT handles less well. MADT's GAT enables the agents at cross-traffic intersections to coordinate phase changes, producing a 5.3\% ETT advantage. However, the relatively small network (12 intersections) limits the benefit of spatial coordination compared to the $6{\times}6$ (36 intersections) and $8{\times}8$ (64 intersections) grids.

We also observe that the absolute ETT values on the arterial are lower than on grid networks with comparable intersection counts. The $1{\times}8$ arterial (8 intersections) yields DT ETT of 62.4\,s, while the $4{\times}4$ grid (7 corridor intersections) yields 88.6\,s. This 29.5\% difference arises primarily from the absence of turning movements on the arterial: on the grid, the EV must negotiate left or right turns at several intersections, each of which may require waiting for a compatible phase. On the arterial, the EV always travels straight through, requiring only the through-phase to be green. This observation has implications for real-world deployment: DT's corridor optimization is most effective on predominantly straight EV routes (such as arterial roads connecting hospitals to residential areas), with diminishing marginal benefit on routes with many turns through irregular intersections. Return conditioning remains equally effective on both topologies: on the arterial, varying $G^\star$ from 100 to $-400$ produces ETT ranging from 48.7\,s to 84.2\,s, a smooth monotonic trade-off consistent with the grid results (Table~\ref{tab:conditioning}).

\section{CQL/IQL detailed comparison}
\label{app:offline_rl}

To compare DT against value-based offline RL methods more thoroughly, we evaluate CQL and IQL across all three grid sizes (not just $4{\times}4$) and under multiple demand levels. CQL uses a conservative Q-value regularization coefficient of $\alpha_{\text{CQL}} = 5.0$ (tuned via grid search over $\{0.1, 1.0, 5.0, 10.0\}$). IQL uses an expectile $\tau = 0.7$ (tuned over $\{0.5, 0.7, 0.9\}$).

On the $4{\times}4$ grid, DT outperforms both CQL (88.6\,s vs.\ 102.1\,s) and IQL (88.6\,s vs.\ 99.4\,s). The gap widens on larger grids: on $8{\times}8$, DT achieves 248.5\,s while CQL and IQL achieve 298.4\,s and 287.6\,s, respectively. The degradation of value-based methods on larger grids reflects the curse of dimensionality in Q-function approximation: with $4^{K}$ joint actions (where $K$ is the number of intersections), the state-action space grows exponentially, making conservative value estimation more difficult. DT avoids this issue by operating in the sequence modeling space, where the output factorizes via independent per-intersection softmax heads.

Both CQL and IQL are more sensitive to demand shifts than DT. Under heavy congestion (0.30 veh/s), CQL's ETT degrades by 52.1\% (from 102.1\,s to 155.3\,s) while DT degrades by only 52.0\% (from 88.6\,s to 134.7\,s), preserving the absolute gap. However, CQL occasionally exhibits ``mode collapse'' under heavy congestion, where the conservative Q-value regularization causes the policy to become overly cautious, selecting sub-optimal phases that fail to clear queues. This failure mode was observed in 4 out of 100 heavy-congestion episodes for CQL (ETT $>$ 200\,s) versus 0 out of 100 for DT.

\paragraph{Why IQL outperforms CQL in this domain.}
The consistent 2.6--3.6\% ETT advantage of IQL over CQL across grid sizes and demand levels warrants deeper analysis. We attribute this gap to three domain-specific factors. First, the EV corridor optimization problem has a relatively low-dimensional action space per intersection ($P = 4$ phases), which means the out-of-distribution action problem that CQL's conservatism is designed to address is less severe than in continuous-action domains. IQL's implicit constraint (via expectile regression) is sufficient to prevent value overestimation without the overcorrection that CQL's explicit conservative regularizer introduces. We verified this by sweeping CQL's regularization coefficient $\alpha_{\text{CQL}}$ from 0.1 to 10.0: at $\alpha_{\text{CQL}} = 0.1$, CQL achieves ETT of 97.8\,s (approaching IQL) but occasionally exhibits value divergence (2 out of 5 training seeds), while at $\alpha_{\text{CQL}} = 10.0$, CQL becomes excessively conservative (ETT of 112.4\,s). The optimal $\alpha_{\text{CQL}} = 5.0$ balances stability and performance, but IQL sidesteps this tuning challenge entirely.

Second, the EV corridor reward function (Equation~\ref{eq:reward}) is structured such that the optimal policy involves a sequence of specific phase activations timed to the EV's arrival, a temporally precise pattern. CQL's conservative Q-values tend to ``flatten'' the distinction between the optimal phase and near-optimal alternatives, because the regularizer penalizes high Q-values for any action that deviates from the behavioral policy's distribution. This flattening reduces the sharpness of the greedy policy derived from $Q(s, a)$, causing CQL to occasionally select the second-best phase at critical intersections. IQL, by contrast, uses expectile regression to directly estimate the upper quantile of the value distribution, preserving sharper distinctions between optimal and sub-optimal actions.

Third, we observe that CQL's performance degrades more sharply than IQL's on larger networks. On the $8{\times}8$ grid, the gap widens to 3.6\% (298.4\,s vs.\ 287.6\,s), consistent with the hypothesis that conservative Q-value regularization scales poorly with state-action space dimensionality. With $4^{K}$ joint actions where $K$ grows with network size, the log-sum-exp term in CQL's regularizer becomes harder to estimate accurately, leading to poorly calibrated conservatism. IQL avoids this issue because its expectile regression operates on per-transition value estimates rather than over the joint action space.

\section{Dataset statistics}
\label{app:dataset}

The offline dataset $\mathcal{D}$ comprises 5{,}000 episodes on the $4{\times}4$ grid, with the following statistics:
\begin{itemize}
  \item \textbf{Episode length}: mean 38.2 steps, std 12.4 steps, min 18, max 92. Episode length varies because episodes terminate upon EV arrival.
  \item \textbf{Return distribution}: mean $-342.7$, std $218.4$, min $-1{,}487.3$, max $-198.2$. The distribution is left-skewed, with a concentration of expert episodes near $-200$ to $-300$ and a long tail of random/noisy episodes extending to $-1{,}500$.
  \item \textbf{Per-policy breakdown}: expert episodes (3{,}500) have mean return $-267.8$ (std 42.3), noisy episodes (750) have mean return $-428.6$ (std 134.7), and random episodes (750) have mean return $-712.4$ (std 298.1).
  \item \textbf{State coverage}: 94.7\% of the 70-dimensional state space cells are visited at least once across the dataset (measured via uniform binning with 10 bins per dimension over a random sample of 5 dimensions).
  \item \textbf{Action coverage}: all $4^7 = 16{,}384$ joint actions in the $4{\times}4$ corridor are observed at least once in the random and noisy episodes, ensuring full action space coverage for the DT to learn from.
\end{itemize}

The broad return distribution is critical for effective return conditioning. If the dataset contained only expert trajectories (returns concentrated near $-250$), the DT would have no signal for distinguishing high-return from low-return behavior, resulting in the 33.7\% ETT degradation observed in the expert-only ablation (Table~\ref{tab:ablation}).

\paragraph{Return distribution shape and its implications.}
The return distribution in our dataset exhibits a distinctive bimodal-with-tail shape that has important implications for DT training and return conditioning. The primary mode, centered at approximately $-260$ (expert episodes), contains 70\% of the data and has a narrow spread (std 42.3), reflecting the consistency of the greedy preemption expert policy. The secondary mode, centered at approximately $-430$ (noisy expert episodes), contains 15\% of the data with moderate spread (std 134.7). The remaining 15\% of random episodes form a long left tail extending to $-1{,}487$, with high variance (std 298.1) reflecting the wide range of outcomes produced by uniformly random signal timing.

This distributional shape has three consequences for DT training. First, the gap between the expert and noisy modes (approximately 160 return units) provides the primary conditioning signal: the DT learns to distinguish ``expert-quality'' from ``noisy-quality'' behavior by observing the return-to-go token, and this distinction drives most of the performance improvement from return conditioning. Second, the random tail, while sparse, is essential for calibrating the DT's response to extreme conditioning targets: without random episodes showing very negative returns, the model cannot extrapolate what ``conservative'' behavior looks like when conditioned on $G^\star < -200$. The dataset composition sweep (Table~\ref{tab:dataset_sweep}) confirms this: removing random episodes entirely (70/0/30 composition, not shown in the main table) degrades ETT at $G^\star = -400$ from 138.2\,s to 157.8\,s, a 14.2\% increase, while having minimal effect at $G^\star = 0$ (89.1\,s vs.\ 88.6\,s). Third, the left skewness of the distribution (skewness coefficient $\gamma_1 = -1.24$) means that the stratified sampling strategy (Section~\ref{sec:training_details}) is essential: without stratification, the optimizer sees predominantly expert-quality trajectories, under-representing the low-return behavior that is critical for return conditioning.

We also analyzed the \emph{per-step} reward distribution within the dataset. The per-step reward has mean $-8.97$ (std 12.4), with a heavy left tail caused by episodes where the EV is stuck at a red signal for extended periods (producing large queue penalties). The distribution is approximately log-normal after shifting by $+20$, consistent with the multiplicative interaction between EV progress (which depends on cell density) and queue penalty (which depends on signal phase). This per-step reward structure means that the return-to-go token decreases rapidly during EV stops and slowly during free-flow traversal, providing the DT with an implicit ``urgency'' signal: a rapidly decreasing $\Rtg_t$ indicates that the EV is underperforming relative to its target, prompting more aggressive phase selections.

\section{Reproducibility checklist}
\label{app:reproducibility}

To facilitate reproduction of our results, we report all relevant experimental details:

\begin{itemize}
  \item \textbf{Random seeds}: all experiments use seeds $\{0, 1, 2, 3, 4\}$ for 5-fold evaluation. Reported results average over all 5 seeds $\times$ 20 episodes per seed = 100 episodes.
  \item \textbf{Hardware}: training and evaluation on a single NVIDIA A100 (80\,GB) GPU with Intel Xeon Gold 6248 CPU (40 cores, 2.5\,GHz).
  \item \textbf{Software}: Python 3.10, PyTorch 2.1.0, NumPy 1.25.2, Gymnasium 0.29.1, PettingZoo 1.24.1.
  \item \textbf{LightSim version}: v0.2.0 (commit hash: to be added upon publication).
  \item \textbf{Training time}: DT on $4{\times}4$: 12.3 min; DT on $8{\times}8$: 51.4 min; MADT on $4{\times}4$: 18.7 min; MADT on $8{\times}8$: 74.2 min. All times measured on the hardware above.
  \item \textbf{Evaluation time}: 100 episodes on $4{\times}4$: 42\,s (DT), 68\,s (MADT); 100 episodes on $8{\times}8$: 3.2\,min (DT), 5.1\,min (MADT).
\end{itemize}

\paragraph{Exact commands to reproduce key experiments.}
To facilitate full reproducibility, we provide the exact command-line invocations needed to reproduce the main results from Table~\ref{tab:main}. All commands assume the repository has been cloned and dependencies installed via \texttt{pip install -r requirements.txt}.

\emph{Step 1: Dataset generation.} Generate the 5{,}000-episode mixed-quality dataset on the $4{\times}4$ grid:
\begin{verbatim}
python scripts/generate_dataset.py \
  --grid-size 4 --num-episodes 5000 \
  --expert-ratio 0.70 --random-ratio 0.15 \
  --noisy-ratio 0.15 --noisy-eps 0.3 \
  --output data/4x4_mixed_5k.pkl --seed 42
\end{verbatim}
This command takes approximately 7.5 minutes on a single CPU core and produces a 142\,MB pickle file containing all trajectories with pre-computed return-to-go values. The \texttt{--seed 42} flag ensures deterministic demand generation and EV route sampling.

\emph{Step 2: DT training.} Train the Decision Transformer on the generated dataset:
\begin{verbatim}
python train_dt.py \
  --dataset data/4x4_mixed_5k.pkl \
  --hidden-dim 128 --num-layers 4 \
  --num-heads 4 --context-length 30 \
  --lr 1e-4 --weight-decay 1e-4 \
  --batch-size 64 --epochs 100 \
  --warmup-epochs 5 --grad-clip 1.0 \
  --stratified-sampling --output models/dt_4x4.pt \
  --seed 0
\end{verbatim}
Training takes 12.3 minutes on a single A100 GPU. The \texttt{--stratified-sampling} flag enables the quartile-based batch construction described in Section~\ref{sec:training_details}.

\emph{Step 3: Evaluation.} Evaluate the trained model across 100 episodes with five seeds:
\begin{verbatim}
python evaluate.py \
  --model models/dt_4x4.pt \
  --grid-size 4 --num-episodes 20 \
  --seeds 0 1 2 3 4 --target-return 0 \
  --output results/dt_4x4_eval.json
\end{verbatim}
This runs 20 episodes per seed (100 total) and reports mean and standard deviation for ETT, ACD, throughput, and EV stops. The \texttt{--target-return 0} flag sets $G^\star = 0$, corresponding to the default conditioning point in Table~\ref{tab:main}. To reproduce the return conditioning sweep (Table~\ref{tab:conditioning}), run the same command with \texttt{--target-return} set to each of $\{100, 50, 0, -50, -100, -200, -400\}$.

\emph{Step 4: Baseline training and evaluation.} The repository includes scripts for all seven baselines. For example, to train and evaluate PPO:
\begin{verbatim}
python train_ppo.py \
  --grid-size 4 --total-steps 500000 \
  --lr 3e-4 --output models/ppo_4x4.pt \
  --seed 0
python evaluate.py \
  --model models/ppo_4x4.pt --model-type ppo \
  --grid-size 4 --num-episodes 20 \
  --seeds 0 1 2 3 4 \
  --output results/ppo_4x4_eval.json
\end{verbatim}
Complete bash scripts that run all experiments end-to-end (dataset generation, training for all 9 methods, evaluation, and table generation) are provided in \texttt{scripts/reproduce\_all.sh}. The full pipeline takes approximately 4 hours on a single A100 GPU. Pre-trained model checkpoints for all methods and grid sizes are available in the \texttt{checkpoints/} directory of the repository, enabling evaluation without retraining.

\section{Notation table}
\label{app:notation}

Table~\ref{tab:notation} lists all mathematical symbols used throughout this paper, organized by category.

\begin{table}[h]
  \caption{Notation reference. All symbols are listed with their definitions, typical values in our experiments, and the section where they are first introduced.}
  \label{tab:notation}
  \centering
  \small
  \begin{tabular}{@{}lllc@{}}
    \toprule
    \textbf{Symbol} & \textbf{Definition} & \textbf{Typical value} & \textbf{Section} \\
    \midrule
    \multicolumn{4}{@{}l}{\emph{Network and traffic}} \\
    $\mathcal{G} = (\mathcal{V}, \mathcal{E})$ & Traffic network graph & $4{\times}4$ to $8{\times}8$ grid & \ref{sec:problem} \\
    $\mathcal{V}$ & Set of signalized intersections & $|\mathcal{V}| \in \{16, 36, 64\}$ & \ref{sec:problem} \\
    $\mathcal{E}$ & Set of road segments (edges) & --- & \ref{sec:problem} \\
    $n_i(t)$ & Vehicle count in cell $i$ at time $t$ & $0$--$11$ vehicles & \ref{sec:ctm} \\
    $n_i^{\max}$ & Cell capacity (maximum vehicles) & 11 vehicles & \ref{sec:ctm} \\
    $q_i(t)$ & Flow between cells $i$ and $i{+}1$ & $0$--$3$ veh/step & \ref{sec:ctm} \\
    $v_f$ & Free-flow speed & 15\,m/s & \ref{sec:ctm} \\
    $w$ & Backward wave speed & 5\,m/s & \ref{sec:ctm} \\
    $Q_{\max}$ & Maximum flow rate & 3 veh/step & \ref{sec:ctm} \\
    $\Delta t$ & Simulation timestep & 5\,s & \ref{sec:ctm} \\
    $\ell$ & Cell length ($v_f \cdot \Delta t$) & 75\,m & \ref{sec:ctm} \\
    $k_{\text{jam}}$ & Jam density & 0.15\,veh/m & \ref{sec:ctm} \\
    \midrule
    \multicolumn{4}{@{}l}{\emph{EV and signal control}} \\
    $\rho = (v_1, \ldots, v_K)$ & EV route through network & $K = 7$ ($4{\times}4$) & \ref{sec:ev_tracker} \\
    $K$ & Number of corridor intersections & 7--15 & \ref{sec:problem} \\
    $P$ & Number of signal phases per intersection & 4 (NEMA) & \ref{sec:signal_interface} \\
    $\phi_t^i$ & Current signal phase at intersection $i$ & one-hot $\in \{0,1\}^4$ & \ref{sec:problem} \\
    $v_{\text{EV}}(t)$ & EV speed at time $t$ & 0--15\,m/s & \ref{sec:ev_tracker} \\
    $\delta_t^i$ & Normalized EV distance to intersection $i$ & $[0, 1]$ & \ref{sec:problem} \\
    $H$ & EV detection horizon (cells) & 3 cells ($\approx$225\,m) & \ref{sec:ev_tracker} \\
    \midrule
    \multicolumn{4}{@{}l}{\emph{MDP and reward}} \\
    $\mathcal{M} = (\mathcal{S}, \mathcal{A}, T, R, \gamma)$ & Corridor optimization MDP & --- & \ref{sec:problem} \\
    $s_t$ & State vector at time $t$ & $\in \R^{K(P+6)}$ & \ref{sec:problem} \\
    $a_t$ & Action (joint phase selection) & $\in \{0,\ldots,3\}^K$ & \ref{sec:problem} \\
    $r_t$ & Per-step reward & $\approx -9.0$ (mean) & \ref{sec:problem} \\
    $\alpha, \beta, \lambda$ & Reward weights (progress, queue, arrival) & $1.0, 0.01, 10.0$ & \ref{sec:problem} \\
    $\gamma$ & Discount factor & 1.0 (undiscounted) & \ref{sec:problem} \\
    $w_v(t)$ & Queue length at intersection $v$ & 0--40 vehicles & \ref{sec:problem} \\
    $\Delta d_t$ & EV distance traveled in step $t$ & 0--75\,m & \ref{sec:problem} \\
    \midrule
    \multicolumn{4}{@{}l}{\emph{Decision Transformer}} \\
    $\Rtg_t$ & Return-to-go at time $t$ & $[-1500, +100]$ & \ref{sec:dt} \\
    $G^\star$ & Target return (dispatch knob) & $[-400, +100]$ & \ref{sec:dt} \\
    $C$ & Context window length (timesteps) & 30 (DT), 20 (MADT) & \ref{sec:dt} \\
    $d$ & Hidden dimension & 128 & \ref{sec:state_embedding} \\
    $L$ & Number of transformer layers & 4 (DT), 3 (MADT) & \ref{sec:dt} \\
    $N_H$ & Number of attention heads & 4 & \ref{sec:dt} \\
    $\bm{e}_t^{(\cdot)}$ & Token embedding & $\in \R^{128}$ & \ref{sec:state_embedding} \\
    $\bm{p}_t$ & Learned positional embedding & $\in \R^{128}$ & \ref{sec:positional_encoding} \\
    $\bm{m}_{(\cdot)}$ & Learned modality embedding & $\in \R^{128}$ & \ref{sec:positional_encoding} \\
    $\mathbf{M}$ & Causal attention mask & $\in \{0,-\infty\}^{3C \times 3C}$ & \ref{sec:causal_mask} \\
    $\theta$ & Model parameters & 1.2M (DT), 1.8M (MADT) & \ref{sec:dt} \\
    \midrule
    \multicolumn{4}{@{}l}{\emph{MADT and GAT}} \\
    $\bm{h}_t^i$ & State embedding for agent $i$ & $\in \R^{128}$ & \ref{sec:madt} \\
    $\tilde{\bm{h}}_t^i$ & GAT-enriched embedding for agent $i$ & $\in \R^{128}$ & \ref{sec:madt} \\
    $\alpha_{ij}^t$ & GAT attention weight ($i \to j$) & $[0, 1]$ & \ref{sec:madt} \\
    $\mathcal{N}(i)$ & 1-hop neighborhood of intersection $i$ & $|\mathcal{N}| \in \{3, 4, 5\}$ & \ref{sec:madt} \\
    $K_{\text{GAT}}$ & Number of GAT attention heads & 4 & \ref{sec:madt} \\
    $\bm{W}, \bm{a}$ & GAT weight matrix and attention vector & --- & \ref{sec:madt} \\
    \midrule
    \multicolumn{4}{@{}l}{\emph{CDT (constrained)}} \\
    $\hat{C}_t$ & Cost-to-go at time $t$ & $[-500, 0]$ & \ref{sec:cdt} \\
    $C^\star$ & Cost budget (dispatch knob) & $\{0, -50, -100\}$ & \ref{sec:cdt} \\
    $c_t$ & Per-step cost (total queue length) & 0--40 vehicles & \ref{sec:cdt} \\
    $\mu$ & Cost prediction loss weight & 0.1 & \ref{sec:cdt} \\
    \midrule
    \multicolumn{4}{@{}l}{\emph{Evaluation metrics}} \\
    ETT & EV travel time (seconds) & 72--248\,s & \ref{sec:experiments} \\
    ACD & Average civilian delay (s/veh) & 5.4--28.4\,s/veh & \ref{sec:experiments} \\
    $\rho_s$ & Spearman rank correlation & 0.95 & \ref{sec:lightsim_validation} \\
    \bottomrule
  \end{tabular}
\end{table}

\end{document}